%% file: main.tex
\documentclass[Afour,sageh,times]{sagej}

\usepackage{bbding}
\usepackage{moreverb,url}
\usepackage[colorlinks,bookmarksopen,bookmarksnumbered,citecolor=blue,urlcolor=blue]{hyperref}
\usepackage{url}
\usepackage{balance}
\usepackage{graphicx}
\usepackage{amsfonts}
\usepackage{fancyhdr}
\usepackage{comment}
\usepackage{times}
\usepackage{amsmath}
\usepackage{changepage}
\usepackage{amssymb}
\usepackage{enumerate}
\usepackage{algorithmic}
\usepackage{bm}
\usepackage{calligra}
\usepackage{multirow}
\usepackage{tabularx}
\usepackage{booktabs}
\usepackage{mathtools}
\usepackage{arydshln}
\usepackage{latexsym} 
\usepackage{amssymb}
\usepackage{color}
\usepackage[ruled,vlined]{algorithm}
\usepackage{ulem}
\usepackage{float}
\usepackage{siunitx}
\usepackage{colortbl}
\usepackage{hhline}
\usepackage{bookmark}
\usepackage{dirtree}
\usepackage{wasysym}
\usepackage{enumitem}
\usepackage{overpic}
\usepackage{pdflscape}
\usepackage{afterpage}

\usepackage{array}
\usepackage{rotating}                                                        
\usepackage{subfig}
\usepackage{textcomp}
\usepackage{xcolor}
\usepackage{booktabs}
\usepackage{multicol}
\usepackage{threeparttable}
\usepackage{lipsum}
\usepackage{color}
\usepackage{makecell}
\usepackage{tikz}
\usetikzlibrary{positioning}
\usepackage{varwidth}
\definecolor{babyblue}{rgb}{0.54, 0.81, 0.94}
\definecolor{bisque}{rgb}{1.0, 0.89, 0.77}
\definecolor{bshade}{rgb}{0.55,0.75,0.95}

\newcommand\BibTeX{{\rmfamily B\kern-.05em \textsc{i\kern-.025em b}\kern-.08em
T\kern-.1667em\lower.7ex\hbox{E}\kern-.125emX}}

\def\volumeyear{2024}

\begin{document}

\runninghead{Chen et al.}

\title{Heterogeneous LiDAR Dataset for Benchmarking Robust Localization in Diverse Degenerate Scenarios}

\author{Zhiqiang Chen\affilnum{1,2},
Yuhua Qi\affilnum{1},
Dapeng Feng\affilnum{1},
Xuebin Zhuang\affilnum{1},
Hongbo Chen\affilnum{1},
Xiangcheng Hu\affilnum{3},
Jin Wu\affilnum{3},
Kelin Peng\affilnum{1},
Peng Lu\affilnum{2}
}

\affiliation{\affilnum{1}Sun Yat-sen University, Guangzhou, China.\\
    \affilnum{2}The University of Hong Kong, Hong Kong SAR, China.\\
    \affilnum{3}The Hong Kong University of Science and Technology, Hong Kong SAR, China
}

\corrauth{Yuhua Qi, School of Systems Science and Engineering, Sun Yat-sen
University, No. 135, Xingang West Road, Guangzhou, China.}
\email{qiyh8@mail.sysu.edu.cn}

\begin{abstract}
    The ability to estimate pose and generate maps using 3D LiDAR significantly enhances robotic system autonomy. However, existing open-source datasets lack representation of geometrically degenerate environments, limiting the development and benchmarking of robust LiDAR SLAM algorithms.
    To address this gap, we introduce GEODE, a comprehensive multi-LiDAR, multi-scenario dataset specifically designed to include real-world geometrically degenerate environments. GEODE comprises 64 trajectories spanning over 64 kilometers across seven diverse settings with varying degrees of degeneracy. The data was meticulously collected to promote the development of versatile algorithms by incorporating various LiDAR sensors, stereo cameras, IMUs, and diverse motion conditions.
    We evaluate state-of-the-art SLAM approaches using the GEODE dataset to highlight current limitations in LiDAR SLAM techniques. This extensive dataset will be publicly available at \href{https://github.com/PengYu-Team/GEODE_dataset}{https://github.com/PengYu-Team/GEODE\_dataset}, supporting further advancements in LiDAR-based SLAM.
\end{abstract}


\maketitle

\input{chapter/introduction}

\input{chapter/related_works}
\input{chapter/system_overview}

\input{chapter/dataset_description}
\input{chapter/experiment}
\section{Known Issues}
\label{sec:known_issue}

Despite our meticulous efforts to construct a multifaceted dataset encompassing diverse platforms, sensors, and scenarios, the process is inherently challenging and not without imperfections. We acknowledge the limitations in our data processing and the persistent challenges despite our diligent data collection and curation efforts. In the following sections, we outline these prevalent issues and propose technical mitigations, hoping to offer future researchers valuable insights and instructive lessons.

\subsection{Camera Exposure Setting}

We set the camera exposure time to a fixed value to reduce image differences caused by lighting changes, making the dataset more uniform. However, this introduces limitations, such as the inability to obtain optimal image quality in extremely bright or dark environments. This issue is particularly evident in underground scenes with insufficient lighting, leading to underexposed images with limited visibility, which poses challenges for most visual perception algorithms.

\subsection{Calibration among Multiple Sensors}

Although each collection device is calibrated using a robust calibration scheme and mounted on rigid aluminum-alloy-based parts to minimize external interference, the calibration process is conducted only once per machine. For data collection efforts spanning a week, the calibration results may not maintain high precision for specific iterations. We recommend users start with our calibration results and investigate new methods for online calibration.

\subsection{Time synchronization}

Our synchronization technique ensures that the LiDAR, cameras, and IMU are all activated by the same clock. However, the computer assigns timestamps to each sensor’s data differently due to varying transmission latencies and decoding times. The LiDAR directly receives GNSS time as its clock source, so delays caused by data transmission and processing are negligible. However, the IMU and camera do not have direct access to GNSS time. For the IMU, we disregard minor transmission and processing delays, synchronizing GNSS time with the computer NUC used for acquisition and running the ROS driver to time the messages. For the camera, we use shared memory to store the latest time and estimate the exposure time to obtain image timestamps. Despite these measures, ensuring perfect synchronization of all sensor data remains challenging. Users requiring higher timestamp accuracy can estimate the time differences between different sensors online.

\section{Conclusion and Future Work}
\label{sec:conclusion}
\runninghead{CHEN \textit{et~al.}}

This study introduces GEODE, a meticulously crafted dataset designed to improve robustness in geometrically degenerate environments. The dataset features three data collection devices, each equipped with different types of LiDARs, as well as the same stereo camera and IMU, all of which have been carefully calibrated and synchronized. These devices are installed on various platforms, collecting data across seven different degenerate scenarios: flat surfaces, stairs, subway tunnels, off-road terrain, inland waterways, urban tunnels, and bridges. GEODE provides a valuable benchmark for LiDAR-based odometry in previously underrepresented scenarios, encouraging a shift in research focus from achieving higher accuracies in geometrically simple cases to enhancing robustness in more complex environments.

Looking ahead, our goal is to extend the dataset to include more degenerate scenarios and a greater variety of robots. A key aspect of this expansion is to incorporate aerial vehicles, enabling collaboration with ground robots in degenerate scenarios for improved positioning. Additionally, we plan to include more sensors and explore new sensor configurations, such as event cameras and advanced FMCW LiDAR, to enhance the dataset's capabilities. Furthermore, we aim to continue collecting new data to cover a broader range of scenes and environments, including extraterrestrial-like settings. By continuously expanding the GEODE dataset, we strive to improve the effectiveness and reliability of robots in diverse real-world situations.

\begin{acks}

    The authors gratefully acknowledge Shipeng Zhong, Ruilan Yang and Yizhen Yin for their contributions to sensor calibration and ground truth pose processing. We also thank the organizational staff and participants for their involvement in the exercises.
\end{acks}

\begin{dci}
    The authors declared no potential conflicts of interest with respect to the research, authorship, and/or publication of this article.
\end{dci}

\normalem  
\bibliographystyle{SageH}
\bibliography{ref.bib}


\end{document}

%% file: chapter/introduction.tex
\section{INTRODUCTION}
\label{sec:introduction}
\runninghead{CHEN \textit{et~al.}}
\subsection{Motivation}

\begin{table*}
    \caption{Comparison With Datasets Containing Geometrically Degenerate Scenarios}        
    \centering{}%
    \resizebox{\textwidth}{!}{
            \begin{threeparttable} 
                \begin{tabular}{lcccccccccccccccc}
                        \toprule 
                        \multirow{3}{*}[-0.3cm]{Dataset} & \multicolumn{6}{c}{Platform} & \multicolumn{4}{c}{Environment} & \multicolumn{5}{c}{Sensors} & \multirow{3}{*}[-0.3cm]{GT Pose}\tabularnewline
                        \cmidrule(lr){2-7} \cmidrule(lr){8-11} \cmidrule(lr){12-16} 
                         & \begin{turn}{-90}
                        \end{turn} &  &  &  &  & \multirow{1}{*}{} & \multirow{2}{*}[-0.3cm]{Scenarios} & \multicolumn{3}{c}{Degradation Type\tnote{2}} & \multirow{2}{*}[-0.3cm]{IMU} & \multicolumn{3}{c}{LiDAR} & \multirow{2}{*}[-0.3cm]{Cam} & \tabularnewline
                        \cmidrule{9-11} \cmidrule{13-15} 
                         & \multirow{1}{*}[0.85cm]{\begin{turn}{-90}
                        H./H.\tnote{1}
                        \end{turn}} & \multirow{1}{*}[0.85cm]{\begin{turn}{-90}
                        UGV
                        \end{turn}} & \multirow{1}{*}[0.85cm]{\begin{turn}{-90}
                        Legged
                        \end{turn}} & \multirow{1}{*}[0.85cm]{\begin{turn}{-90}
                        Vehicle
                        \end{turn}} & \multirow{1}{*}[0.85cm]{\begin{turn}{-90}
                        Boat
                        \end{turn}} & \multirow{1}{*}[0.85cm]{\begin{turn}{-90}
                        UAV
                        \end{turn}} &  & Trans. & Rot. & Mixed &  & \# Spinning & \# Non-repetitive & \makecell{\#additional\\channels\tnote{3}} &  & \tabularnewline
                        \midrule 
                        \makecell[l]{UrbanNav\\ \emph{Tunnel Sequence} \\ \cite{Hsu2023HongKU}} & \textcolor{red}{\XSolidBrush{}} & \textcolor{red}{\XSolidBrush{}} & \textcolor{red}{\XSolidBrush{}} & \textcolor{green}{\CheckmarkBold{}} & \textcolor{red}{\XSolidBrush{}} & \textcolor{red}{\XSolidBrush{}} & Tunnel & \textcolor{green}{\CheckmarkBold{}} & \textcolor{red}{\XSolidBrush{}} & \textcolor{red}{\XSolidBrush{}} & \textcolor{green}{\CheckmarkBold{}} & 3 & \textcolor{red}{\XSolidBrush{}} & \textcolor{red}{\XSolidBrush{}} & \textcolor{green}{\CheckmarkBold{}} & GNSS-RTK/INS\tabularnewline
                        \makecell[l]{ECMD\\ \emph{Tunnel} / \emph{Bridge Sequences}\\ \cite{Chen2023ECMDAE}} & \textcolor{red}{\XSolidBrush{}} & \textcolor{red}{\XSolidBrush{}} & \textcolor{red}{\XSolidBrush{}} & \textcolor{green}{\CheckmarkBold{}} & \textcolor{red}{\XSolidBrush{}} & \textcolor{red}{\XSolidBrush{}} & Tunnel / Bridge & \textcolor{green}{\CheckmarkBold{}} & \textcolor{red}{\XSolidBrush{}} & \textcolor{red}{\XSolidBrush{}} & \textcolor{green}{\CheckmarkBold{}} & 3 & \textcolor{red}{\XSolidBrush{}} & \textcolor{red}{\XSolidBrush{}} & \textcolor{green}{\CheckmarkBold{}} & GNSS-RTK/INS\tabularnewline
                        \midrule 
                        \makecell[l]{WHU-Helmet\\ \emph{Underground Sequences}\\  \cite{Li2023WHUHelmetAH}} & \textcolor{green}{\CheckmarkBold{}} & \textcolor{red}{\XSolidBrush{}} & \textcolor{red}{\XSolidBrush{}} & \textcolor{red}{\XSolidBrush{}} & \textcolor{red}{\XSolidBrush{}} & \textcolor{red}{\XSolidBrush{}} & Underground & \textcolor{green}{\CheckmarkBold{}} & \textcolor{red}{\XSolidBrush{}} & \textcolor{red}{\XSolidBrush{}} & \textcolor{green}{\CheckmarkBold{}} & \textcolor{red}{\XSolidBrush{}} & 2 & \textcolor{red}{\XSolidBrush{}} & \textcolor{green}{\CheckmarkBold{}} & SLAM\tabularnewline
                        \makecell[l]{ENWIDE \\ \cite{Pfreundschuh2023COINLIOCI}} & \textcolor{green}{\CheckmarkBold{}} & \textcolor{red}{\XSolidBrush{}} & \textcolor{red}{\XSolidBrush{}} & \textcolor{red}{\XSolidBrush{}} & \textcolor{red}{\XSolidBrush{}} & \textcolor{red}{\XSolidBrush{}} & \makecell{Tunnel / Intersection\\ Runway} & \textcolor{green}{\CheckmarkBold{}} & \textcolor{green}{\CheckmarkBold{}} & \textcolor{green}{\CheckmarkBold{}} & \textcolor{green}{\CheckmarkBold{}} & 1 & \textcolor{red}{\XSolidBrush{}} & \textcolor{green}{\CheckmarkBold{}} & \textcolor{red}{\XSolidBrush{}} & Laser Tracker\tabularnewline
                        \midrule
                        \makecell[l]{CERBERUS\\ \cite{Tranzatto2022CERBERUSIT}} & \textcolor{red}{\XSolidBrush{}} & \textcolor{red}{\XSolidBrush{}} & \textcolor{green}{\CheckmarkBold{}} & \textcolor{red}{\XSolidBrush{}} & \textcolor{red}{\XSolidBrush{}} & \textcolor{red}{\XSolidBrush{}} & Cavern & \textcolor{green}{\CheckmarkBold{}} & \textcolor{red}{\XSolidBrush{}} & \textcolor{red}{\XSolidBrush{}} & \textcolor{green}{\CheckmarkBold{}} & 3 & \textcolor{red}{\XSolidBrush{}} & \textcolor{green}{\CheckmarkBold{}} & \textcolor{green}{\CheckmarkBold{}} & ICP Registration\tabularnewline
                        \makecell[l]{CoSTAR \\ \cite{Chang2022LAMP2A}} & \textcolor{red}{\XSolidBrush{}} & \textcolor{green}{\CheckmarkBold{}} & \textcolor{green}{\CheckmarkBold{}} & \textcolor{red}{\XSolidBrush{}} & \textcolor{red}{\XSolidBrush{}} & \textcolor{red}{\XSolidBrush{}} & Tunnel / Cavern & \textcolor{green}{\CheckmarkBold{}} & \textcolor{red}{\XSolidBrush{}} & \textcolor{red}{\XSolidBrush{}} & \textcolor{red}{\XSolidBrush{}} & 1 & \textcolor{red}{\XSolidBrush{}} & \textcolor{red}{\XSolidBrush{}} & \textcolor{red}{\XSolidBrush{}} & ICP Registration\tabularnewline
                        \makecell[l]{CTU-CRAS-Norlab \\ \cite{Petrek2021LargeScaleEO}}& \textcolor{red}{\XSolidBrush{}} & \textcolor{red}{\XSolidBrush{}} & \textcolor{red}{\XSolidBrush{}} & \textcolor{red}{\XSolidBrush{}} & \textcolor{red}{\XSolidBrush{}} & \textcolor{green}{\CheckmarkBold{}} & Tunnel / Cavern & \textcolor{green}{\CheckmarkBold{}} & \textcolor{red}{\XSolidBrush{}} & \textcolor{red}{\XSolidBrush{}} & \textcolor{green}{\CheckmarkBold{}} & 2 & \textcolor{red}{\XSolidBrush{}} & \textcolor{green}{\CheckmarkBold{}} & \textcolor{green}{\CheckmarkBold{}} & ICP Registration\tabularnewline
                        \makecell[l]{SubT-MRS\\ \cite{zhao2024subt}} & \textcolor{green}{\CheckmarkBold{}} & \textcolor{green}{\CheckmarkBold{}} & \textcolor{green}{\CheckmarkBold{}} & \textcolor{red}{\XSolidBrush{}} & \textcolor{red}{\XSolidBrush{}} & \textcolor{green}{\CheckmarkBold{}} & \makecell{Offroad / Corridor\\ Tunnel / Cavern} & \textcolor{green}{\CheckmarkBold{}} & \textcolor{green}{\CheckmarkBold{}} & \textcolor{green}{\CheckmarkBold{}} & \textcolor{green}{\CheckmarkBold{}} & 1 & \textcolor{red}{\XSolidBrush{}} & \textcolor{red}{\XSolidBrush{}} & \textcolor{green}{\CheckmarkBold{}} & GICP Registration\tabularnewline
                        \midrule
                        Ours & \textcolor{green}{\CheckmarkBold{}} & \textcolor{green}{\CheckmarkBold{}} & \textcolor{red}{\XSolidBrush{}} & \textcolor{green}{\CheckmarkBold{}} & \textcolor{green}{\CheckmarkBold{}} & \textcolor{red}{\XSolidBrush{}} & \makecell{Urban Tunnel / Bridge \\ Flat Ground / Stairs\\  Metro tunnel / Offroad \\Inland Waterways} & \textcolor{green}{\CheckmarkBold{}} & \textcolor{green}{\CheckmarkBold{}} & \textcolor{green}{\CheckmarkBold{}} & \textcolor{green}{\CheckmarkBold{}} & 2 & 1 & \textcolor{green}{\CheckmarkBold{}} & \textcolor{green}{\CheckmarkBold{}} & \makecell{GNSS-RTK/INS\\ Laser Tracker\\ SLAM using GT Map}\tabularnewline
                        \bottomrule
                        \end{tabular}
            \begin{tablenotes}  
                \item[1] H./H. denotes two distinct methods for mounting data collection devices: handheld and helmet-mounted.
                \item[2] We categorize the degenerate scenarios into three distinct situations: degradation manifesting solely in the translational direction, degradation occurring exclusively in the rotational dimension, and degradation occurring concurrently in both translational and rotational directions.
                \item[3] The LiDAR sensor employed possesses the capability to capture supplementary data channels, such as reflectivity information.
            \end{tablenotes}
            \end{threeparttable}
    }
    \label{table:GDS-dataset}
\end{table*}

The demand for robots capable of operating in real-world environments, such as automated mining and disaster response, has been steadily increasing. Despite significant advancements in LiDAR SLAM for autonomous navigation, practical performance remains hindered by errors arising from insufficient geometric constraints in challenging scenarios like tunnels and subways. In these degenerate environments, geometric constraints in certain directions can be indistinguishable from noise, leading state optimization to converge on a noise-induced optimum, known as a degenerate solution \citep{Zhang2016OnDO,Tuna2024}.

The choice of LiDAR hardware significantly influences LiDAR degeneracy. Degeneracy is more prevalent in spinning LiDARs with fewer scan lines, which may lose sight of structural features during movement, compared to those with a higher number of scan lines. Additionally, LiDARs equipped with channels such as reflectivity can enhance localization by leveraging additional environmental modalities, rather than relying solely on spatial features \citep{Zhang2023RILIORI,Pfreundschuh2023COINLIOCI}. Although non-repetitive LiDARs generate denser point clouds than conventional spinning LiDARs, their restricted field of view can lead to a greater loss of spatial features compared to omnidirectional LiDAR sensors, necessitating the deployment of multiple LiDARs \citep{jung2023asynchronous}. There is a notable deficiency in comprehensive datasets that encompass the wide range of LiDAR configurations for various degenerate scenarios, highlighting a gap in the availability of benchmark datasets for validating robust localization with heterogeneous LiDARs.

Current open-source SLAM datasets are predominantly recorded in feature-rich environments such as campuses, parks, or streets, and typically exclude various types of degenerate scenes \citep{Geiger2012AreWR,Liao2021KITTI360AN,Helmberger2021TheHS,Zhang2022HiltiOxfordDA,Feng2022S3EAM}. Only a few datasets \citep{Chang2022LAMP2A,Pfreundschuh2023COINLIOCI,Tranzatto2022CERBERUSIT,zhao2024subt} are conducive to studying LiDAR degeneracy, and these have three main limitations. First, they are often acquired in limited scenarios and lack real-world environments with varying degrees of degeneracy, ranging from mild to severe, in both rotational and translational directions. This hinders the development of refined degeneracy metrics for accurately detecting degradation levels. Second, while some datasets provide multiple degeneracy scenarios, they are not scalable to encompass a range of environmental scales, which is necessary for demonstrating promising performance across different operational contexts. Third, existing datasets mainly focus on a single type of LiDAR without heterogeneous setups. Incorporating a diverse range of LiDAR sensors is essential to investigate versatile methodologies in challenging environments and ensure compatibility with various hardware configurations.

The lack of datasets that fully characterize degenerate scenes complicates accurate localization in environments with perceptual aliasing. Evaluating algorithms on datasets that do not represent the broad spectrum of LiDAR configurations may result in suboptimal performance across hardware setups. These limitations pose significant challenges for state-of-the-art LiDAR SLAM methods when applied to real-world tasks in geometrically degenerate environments. Consequently, creating a dataset that captures and simulates the challenges faced by SLAM algorithms in real-world degenerate scenes using heterogeneous LiDARs is crucial.

\subsection{Contributions}
We have identified a significant deficiency in existing 3D LiDAR datasets, specifically in addressing scenarios characterized by geometrically degenerate conditions. To address this gap, we have developed the GEODE dataset, which includes seven real-world environments with extensive segments featuring \textbf{GEO}metric \textbf{DE}generacies. This is the first publicly available dataset that integrates multiple LiDARs, various scenarios, and is specifically designed to highlight LiDAR degeneracies, complete with accurate ground truth data.
Our contributions are as follows:

First, we designed three devices equipped with different LiDAR types, scanning patterns, scan lines, and fields of view (FOVs), along with stereo cameras and an inertial measurement unit (IMU) for multi-sensor integration. These devices were mounted on various platforms to collect data in highly challenging scenarios, testing LiDAR-based SLAM systems against insufficient geometric features, unpredictable motion patterns, and environmental changes.

Second, the dataset encompasses a range of geometrically degenerate environments, including flat ground, stairs, metro tunnels, off-road terrain, inland waterways, urban tunnels, and bridges. It covers a wide array of LiDAR degradation scenarios, from indoor rooms to highways, and includes sequences that capture environmental changes across different degradation levels. This diversity facilitates the development of robust algorithms for detecting and mitigating geometric degeneracies.

Third, we provide precise ground-truth poses for each sequence and ground-truth maps for select indoor sequences. Our aim is to evaluate state-of-the-art SLAM systems, including four LiDAR-inertial odometry methods and three multi-sensor fusion approaches, to identify the limitations of current LiDAR-centric SLAM algorithms. The GEODE dataset will be publicly released, representing a pioneering large-scale dataset focused on scenes with geometric degeneracies and diverse sensors, thereby pushing the boundaries of LiDAR SLAM research.

\subsection{Organization}
The structure of this paper is as follows: Section \ref{sec:related_works} reviews existing datasets that incorporate heterogeneous LiDAR sensors and geometrically degenerate scenarios, highlighting our key contributions. Section \ref{sec:system_overview} describes the hardware setup and sensor specifications for data collection and ground-truth acquisition, including sensor calibration procedures. Section \ref{sec:dataset_description} provides an overview of the dataset, detailing the scenarios, key features, organization, and ground-truth generation. Section \ref{sec:experiment} outlines the methods used for evaluating localization, highlights the limitations of current LiDAR SLAM systems, and discusses potential solutions for improving robustness. Section \ref{sec:known_issue} addresses the known issues of the GEODE dataset. 
Finally, Section \ref{sec:conclusion} wraps up the paper and proposes avenues for future research.

%% file: chapter/related_works.tex
\section{RELATED WORK}
\label{sec:related_works}
\runninghead{CHEN \textit{et~al.}}

Datasets like the KITTI Dataset \citep{Geiger2012AreWR,Liao2021KITTI360AN} and the NCLT Dataset \citep{carlevaris2016university} have integrated LiDAR technology to enhance the accuracy and reliability of state estimation methods. These datasets facilitate the development and comparison of various LiDAR SLAM solutions, driving continuous improvements in algorithms and localization capabilities. LiDAR-based SLAM methods, which utilize point cloud registration, have shown significant progress with the support of open-source datasets. Techniques such as LOAM \citep{Zhang2014LOAMLO} and Traj-LO \citep{Zheng2023TrajLOID} have demonstrated outstanding performance in the KITTI odometry benchmark. However, these datasets are limited to single LiDAR usage in well-structured scenarios, which restricts their effectiveness in evaluating algorithms for heterogeneous LiDAR sensor arrays and degeneracy-aware localization.

\textbf{Datasets with Heterogeneous LiDAR Sensors.} 
The Pohang Canal dataset \citep{chung2023pohang} employs three spinning LiDARs to improve autonomous navigation in narrow waterways. Similarly, the NTU VIRAL dataset \citep{nguyen2022ntu} uses two spinning LiDARs for localization from an aerial vehicle perspective. Most of these datasets mainly depend on spinning LiDARs. The TIERS dataset \citep{qingqing2022multi}, which includes three spinning LiDARs and three non-repetitive LiDAR sensors for multi-LiDAR SLAM, was introduced to address this limitation. The HILTI series \citep{Helmberger2021TheHS,Zhang2022HiltiOxfordDA} also considers heterogeneous LiDARs for precise LiDAR SLAM, with methods like LiDAR Bundle Adjustment \citep{Liu2022LargeScaleLC} excelling in the HILTI benchmark competition. Additionally, the city dataset \citep{jung2023asynchronous} has been used to evaluate LiDAR SLAM with three heterogeneous LiDAR sensors. Beyond SLAM tasks, heterogeneous LiDAR data is utilized for other applications. For instance, HeLiPR dataset \citep{jung2023helipr}, which includes various types of LiDARs with different scanning patterns and additional measurement channels, is designed for place recognition with heterogeneous LiDARs, capturing spatiotemporal variations. HeLiMOS dataset \citep{lim2024helimos} is a dataset for moving object segmentation in 3D point clouds from heterogeneous LiDAR sensors. Furthermore, a Multi-LiDAR Multi-UAV Dataset \citep{catalano2023towards} was proposed to advance UAV tracking techniques.

Despite the use of heterogeneous LiDARs in these datasets, their primary focus remains on SLAM in well-structured environments, place recognition, or object detection and tracking. There is still a need for datasets that address degradation detection and precise localization using heterogeneous LiDARs in degraded environments.

\begin{table}[t]
    \centering
    \caption{Sensors Specifications and Tracking Devices}
    \label{table:SENSORS-SPECIFICATIONS}
    \resizebox{0.5\textwidth}{!}{
    \begin{tabular}{lll}
    \toprule %
    \multirow{1}{*}{Sensor Type} & \multirow{1}{*}{Type} & \multirow{1}{*}{Unit} \\
    \midrule
    LiDAR $\alpha$ & Velodyne	\\
    \quad{}Model & VLP-16 	\\
    \quad{}Scan lines & 16 \\
    \quad{}Range & 120 & $m$ \\
    \quad{}Vertical FOV & 30 & $deg$\\
    \quad{}Horizontal FOV & 360 & $deg$ \\
    \quad{}Frequency 	& 10 & $Hz$ \\
    LiDAR $\beta$ & Ouster	\\
    \quad{}Model & OS1-64 	\\
    \quad{}Scan lines & 64 \\
    \quad{}Range & 100 & $m$ \\
    \quad{}Vertical FOV & 45 & $deg$\\
    \quad{}Horizontal FOV & 360 & $deg$ \\
    \quad{}Frequency 	& 10 & $Hz$ \\
    \quad{}IMU & InvenSense ICM-20948@100Hz\\
    LiDAR $\gamma$ & Livox	\\
    \quad{}Model & AVIA 	\\
    \quad{}Range & 450 & $m$ \\
    \quad{}Vertical FOV & 30 & $deg$\\
    \quad{}Horizontal FOV & 360 & $deg$ \\
    \quad{}Frequency 	& 10 & $Hz$ \\
    \quad{}IMU & BMI088@200Hz\\
    \midrule
    IMU & Xsens		\\
    \quad{}Model & MTi-30 AHRS \\
    \quad{}Frequency & 100 & $Hz$ \\
    \quad{}Gyro noise density & 0.03 & $^{o}/s/\sqrt{Hz}$\\
    \quad{}Accel noise density & 60 & $\mu g/\sqrt{Hz}$\\ 
    \quad{}Mag RMS noise & 0.5 & $mGauss$\\
    \midrule
    Camera - Stereo & HikRobot - GigE cameras	\\
    \quad{}Model & MV-CS050-10GC \\
    \quad{}Resolution  & 1224 $\times$ 1024 & $pixel$\\
    \quad{}Frequency 	& 10 & $Hz$ \\
    \quad{}Readout Method & Global shutter \\
    \midrule
    GNSS-RTK/INS & CHCNAV	\\
    \quad{}Mode & CGI610 \\
    \quad{}Position Output & NMEA \\
    \quad{}Frequency & 100 & $Hz$\\
    \quad{}RTK Accuracy & 1& $cm$\\
    \bottomrule
    Mocap System & Vicon\\
    \quad{}Mode &  Vero 2.2 \\
    \quad{}Max Frame Rate & 330  &  $Hz$\\
    \quad{}Accuracy & 1 & $mm$\\
    \bottomrule
    Laser Tracker & Leica \\
    \quad{}Mode &  Nova MS60 \\
    \quad{}Frequency & 10 &  $Hz$\\
    \quad{}Accuracy & 1 & $mm$\\
    \bottomrule
    3D Laser Scanner & Leica \\
    \quad{}Mode &  RTC360 \\
    \quad{}Range & 130 & $m$ \\
    \quad{}Vertical FOV & 300 & $deg$\\
    \quad{}Horizontal FOV & 360 & $deg$ \\
    \quad{}Accuracy & 1 & $mm$\\
    \bottomrule
    \end{tabular}
    }
    \label{sen_specs}
\end{table}

\begin{figure*}[t]
    \centering
    \subfloat[Multi-sensor device]{\includegraphics[width=0.47\textwidth]{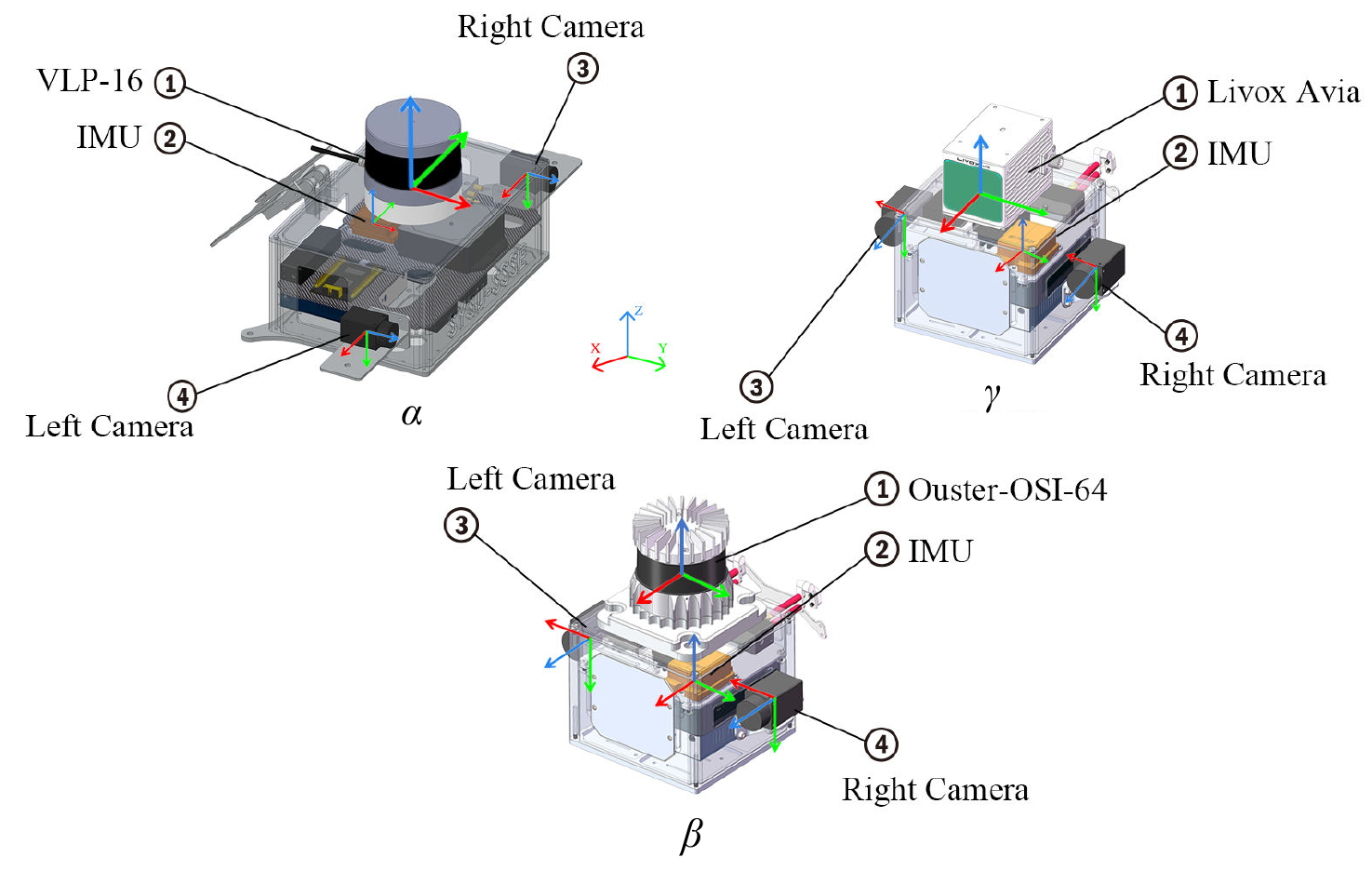}
    \label{fig:Device}}
    \subfloat[Handheld]{\includegraphics[width=0.16\textwidth]{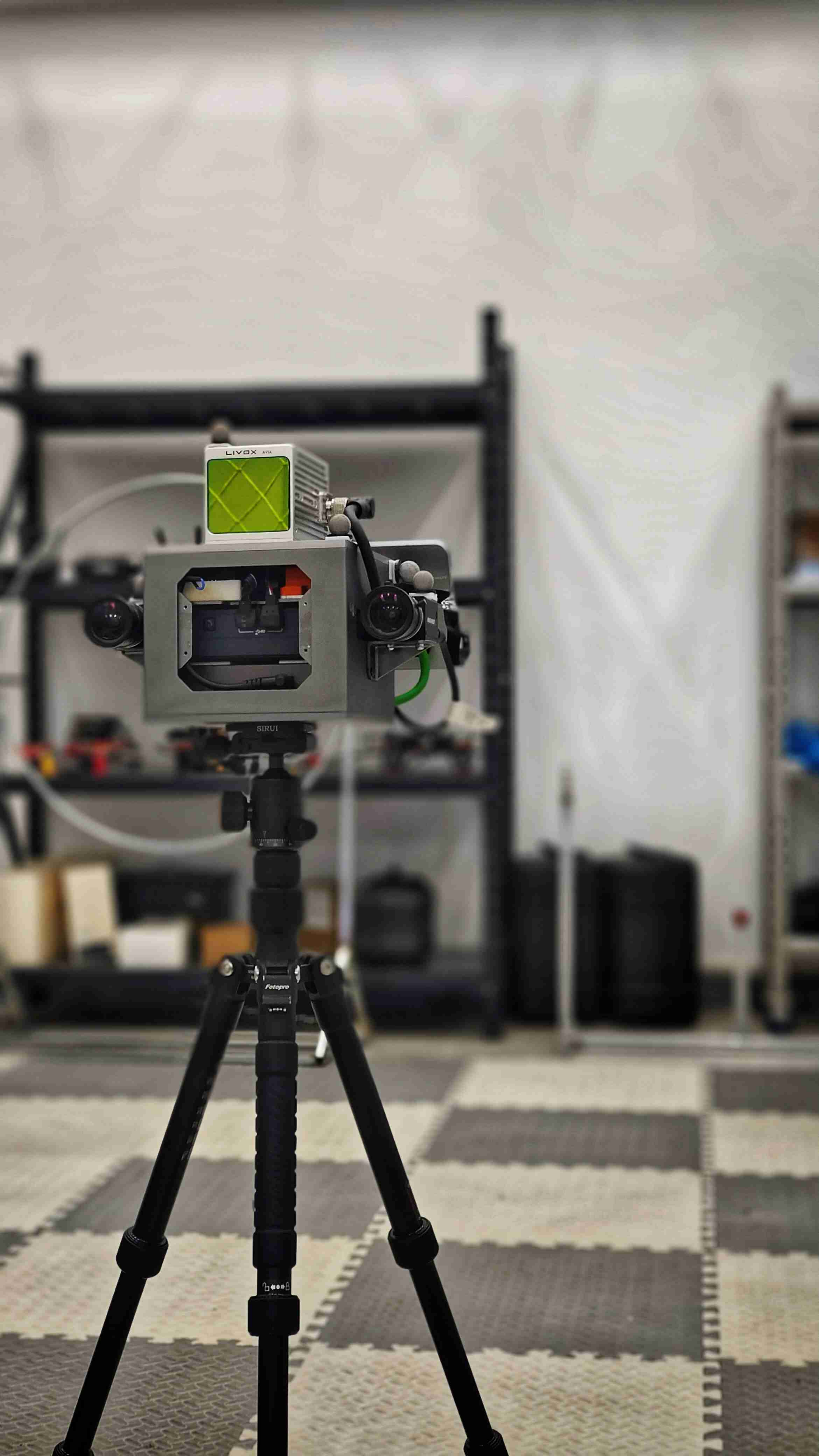}
    \label{fig:Handheld}}
    \subfloat[Sailboat]{\includegraphics[width=0.16\textwidth]{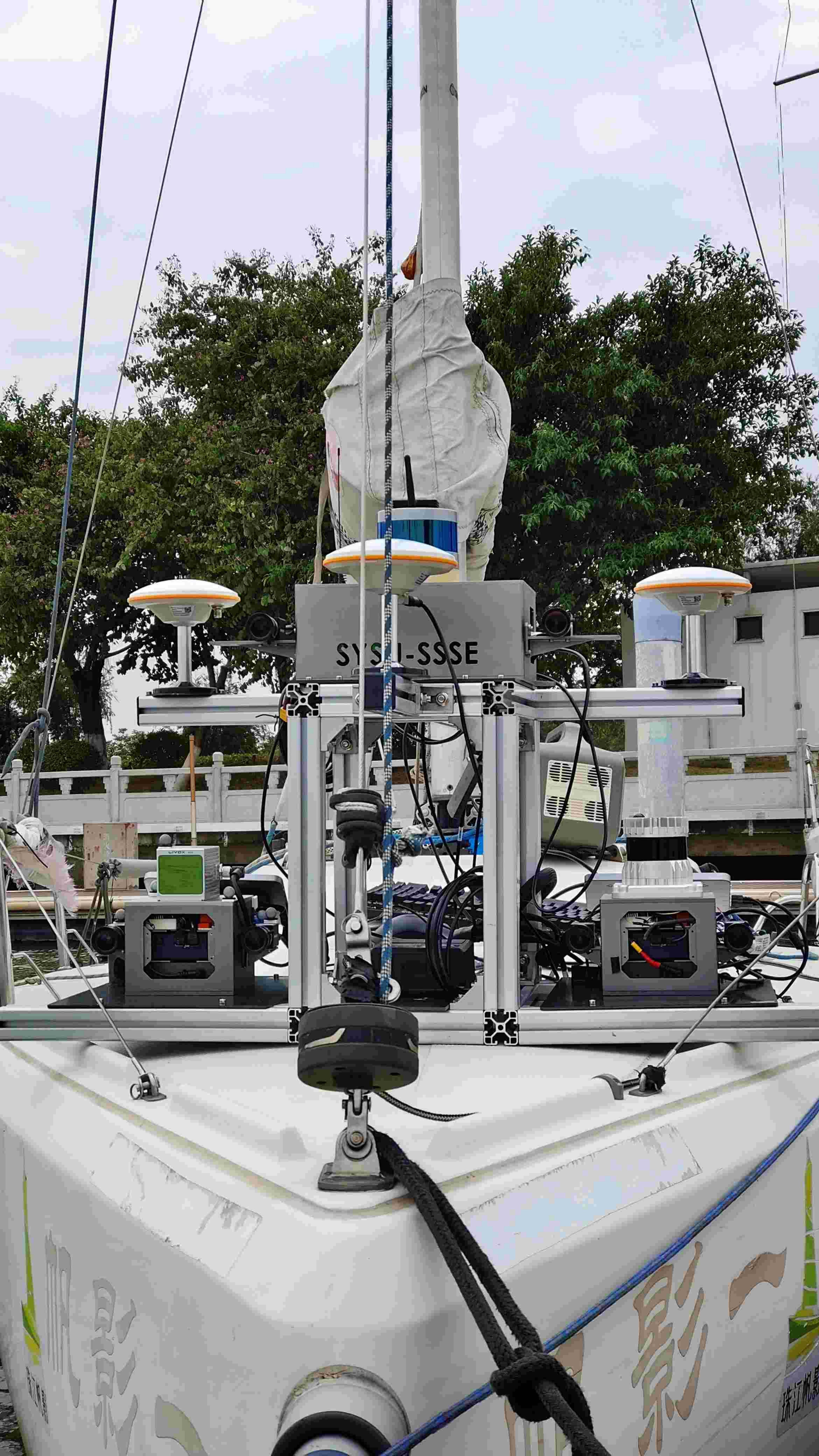}
    \label{fig:Sailboat}}
    \subfloat[UGV]{\includegraphics[width=0.16\textwidth]{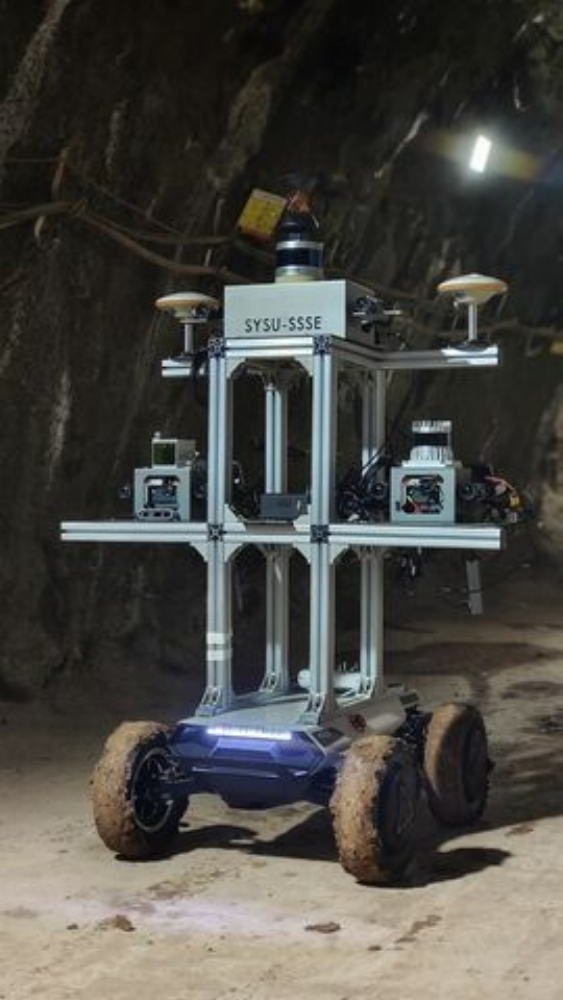}
    \label{fig:UGV}}

    \caption{Multi-Sensor Devices and Data Collection Platforms.
    \textbf{(a)} SolidWorks models of our sensor rig on three data collection devices, with the coordinate axes color-coded: \textcolor{red}{red} for the $X$-axis, \textcolor{green}{green} for the $Y$-axis, and \textcolor{blue}{blue} for the $Z$-axis. This representation illustrates the transformation of sensor coordinates for each device. The multi-sensor rig mounted on
    \textbf{(b)} a handheld platform,
    \textbf{(c)} a sailboat, and
    \textbf{(d)} an UGV.
    The images in (b) through (d) demonstrate the diverse range of the GEODE dataset across various data collection platforms.}
    \label{fig:Multisensor_device}
\end{figure*}

\textbf{Datasets with Multi Degenerate Scenarios.} 
Although several datasets with degenerate scenarios contribute to research on degeneracy-aware localization, each has unique limitations. Table \ref{table:GDS-dataset} compares existing datasets with degenerate environments with our work. While the UrbanNav dataset \citep{Hsu2023HongKU} and ECMD dataset \citep{Chen2023ECMDAE} include degenerate scenarios such as tunnels or bridges in specific sequences, they lack multiple motion kinematic profiles and diverse degenerative scenarios, and a variety of LiDAR suits. The WHU-Helmet dataset \citep{Li2023WHUHelmetAH} and ENWIDE dataset \citep{Pfreundschuh2023COINLIOCI} aim to enhance the robustness of LiDAR SLAM algorithms, focusing on challenging GNSS-denied environments and geometrically degenerate scenarios, respectively. However, these datasets lack hardware diversity, primarily utilizing a single type of LiDAR. The DARPA challenge \citep{Ebadi2024PresentAF} has significantly advanced robust odometry techniques, and participating teams have open-sourced relevant datasets. Nevertheless, the scenarios captured were in limited settings with relatively mild degradation, such as caves, with restricted types of LiDAR sensors and platforms used for data collection. The SubT-MRS dataset \citep{zhao2024subt} by Team Explorer extends this progress by introducing additional scenes with varying degrees of degradation and weather changes collected by multi-robot teams. However, this dataset contains only a single spinning LiDAR, posing a challenge for algorithms aiming to achieve broader hardware compatibility. 

In summary, our dataset exhibits enhanced comprehensiveness in four key aspects: 1) Extensive sensory measurements derived from diverse LiDAR, providing additional channels for LiDAR data; 2) Inclusion of multiple scenarios encompassing varying levels of degradation, enhancing the performance of degeneracy detection and mitigation; 3) Comprehensive data collection incorporating a wide range of motion patterns to facilitate algorithm design for general-purpose applications; and 4) Simulation of potential real-world sensor failures to improve adaptive algorithm switching and failure detection and recovery.

%% file: chapter/system_overview.tex
\section{SYSTEM OVERVIEW}
\label{sec:system_overview}
\runninghead{CHEN \textit{et~al.}}

\subsection{Sensors Setup}
The dataset's design aims to facilitate the development of LiDAR SLAM algorithms, independent of scanning modalities and FoV characteristics. To achieve this objective, we have developed three acquisition devices that share a common IMU and stereo camera but are equipped with distinct LiDAR sensors. The sensor parameters and layout are detailed in Table \ref{table:SENSORS-SPECIFICATIONS} and Figure \ref{fig:Device}. Our versatile acquisition system can be easily mounted on various platforms, as demonstrated in Figure \ref{fig:Handheld},\ref{fig:Sailboat},\ref{fig:UGV}, showcasing its adaptability to a handheld, sailboat, and UGV device, respectively.

\subsection{Time Synchronization Scheme}
\begin{figure}[t]
  \centering
  \includegraphics[width=0.49\textwidth]{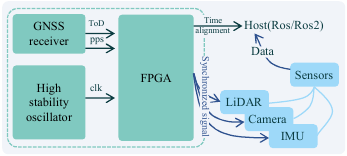}
  \caption{Time synchronization scheme.}
  \label{fig:Time_Synchronization}
\end{figure}

Our FPGA-based synchronization module facilitates multi-channel sensor synchronization, as illustrated in Figure \ref{fig:Time_Synchronization}. The module is capable of achieving outdoor time synchronization through the reception of GNSS signals during initialization. By utilizing TIME\_OF\_DAY (TOD) and PPS signals from the GNSS, it generates synchronized signals at frequencies of 1, 10, and 100 Hz for LiDAR, cameras, and the Xsens MTi-30 IMU, respectively. The embedded IMUs in the Livox Avia and Ouster LiDARs autonomously coordinate with other sensors. In environments where GPS signal is unavailable, accurate synchronization is maintained using an internal clock mechanism.

\subsection{Sensor Calibration}
\label{subsec:Sensor_Calibration} 

\begin{table*}[t]
    \caption{An Overview of Scenarios in GEODE Dataset}
    \centering{}%
    \resizebox{\textwidth}{!}{
    \begin{tabular}{cccccccc}
    \toprule 
    Scenario & Flat Ground & Stairs & Metro Tunnels & Offroad & Inland Waterways & Urban Tunnel & Bridges\tabularnewline
  \midrule
  Number of sequences & 2 & 3 & 23 & 21 & 9 & 3 & 3\tabularnewline
  Size/GB & 1.5 & 17.6 & 153.1 & 152.0 & 147.3 & 10.6 & 12.3\tabularnewline
  Duration/s & 170 & 1066 & 6615 & 8112 & 7436 & 961 & 1174\tabularnewline
  Dist/m & 108.29 & 901.81 & 7524.62 & 12829.20 & 15868.59 & 12975.32 & 14324.33\tabularnewline
  Degeneracy Direction & 2 Trans. + 1 Rot. & 1 Trans. & 1 Trans. + 1 Rot. & 2 Trans. + 1 Rot. & 1 Trans. & 1 Trans. & 1 Trans.\tabularnewline
  Devices(Platform) & $\gamma$(Handheld) & \multicolumn{1}{c}{\begin{tabular}[c]{@{}c@{}}$\alpha$(Handheld)\\$\beta$(Handheld)\\$\gamma$(Handheld)\end{tabular}} & \multicolumn{1}{c}{\begin{tabular}[c]{@{}c@{}}$\alpha$(UGV)\\$\beta$(UGV/Handheld)\\$\gamma$(UGV/Handheld)\end{tabular}} & \multicolumn{1}{c}{\begin{tabular}[c]{@{}c@{}}$\alpha$(UGV)\\$\beta$(UGV)\\$\gamma$(UGV)\end{tabular}} & \multicolumn{1}{c}{\begin{tabular}[c]{@{}c@{}}$\alpha$(Sailboat)\\$\beta$(Sailboat)\\$\gamma$(Sailboat)\end{tabular}} & $\alpha$(Vehicle) & $\alpha$(Vehicle)\tabularnewline
  Loop Closure & \CheckmarkBold{} & \CheckmarkBold{} & \CheckmarkBold{} & \CheckmarkBold{} & \CheckmarkBold{} & \XSolidBrush{} & \XSolidBrush{}\tabularnewline
  GT & Mocap & PALoc & Laser Tracker & GNSS-RTK/INS & GNSS-RTK/INS & GNSS-RTK/INS & GNSS-RTK/INS\tabularnewline
    \bottomrule
    \end{tabular}
    }
    \label{tab:overview_of_dataset}
\end{table*}

\begin{figure*}[t] \centering
  \newcommand{\hwidth}{1pt}
  \makebox[0.24\textwidth]{%
      \includegraphics[width=0.12\textwidth]{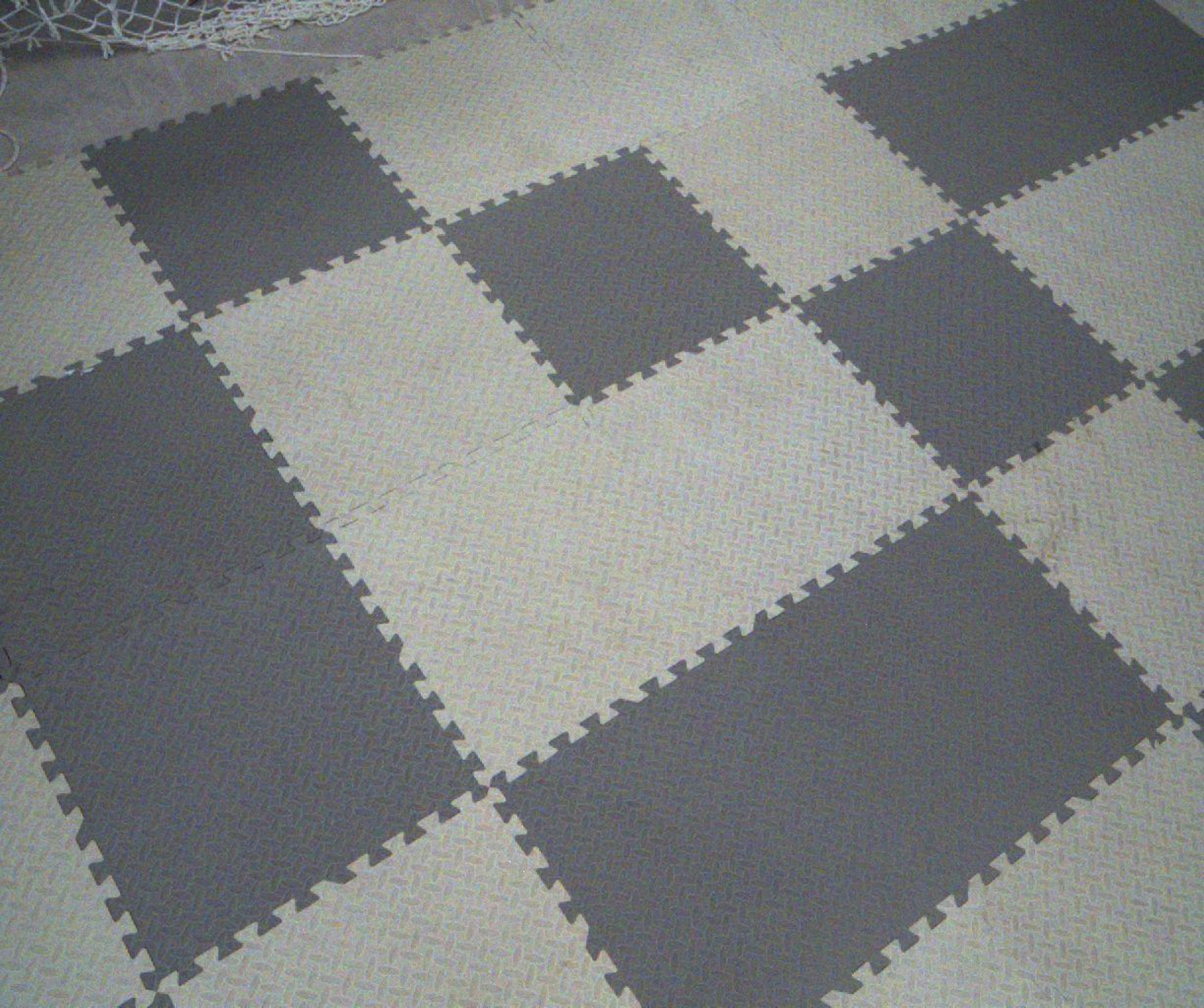}\hspace{\hwidth}
      \includegraphics[width=0.12\textwidth]{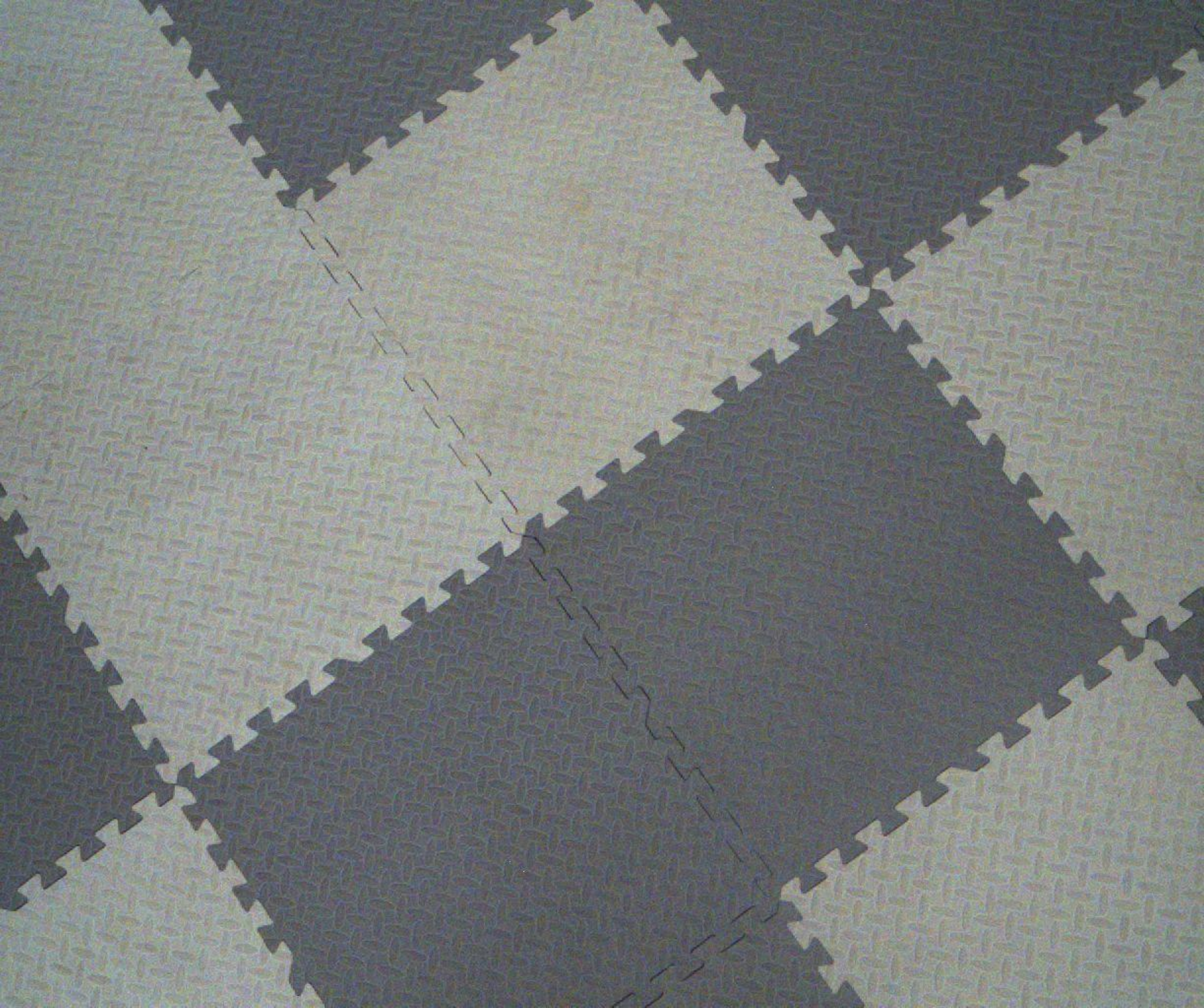}} 
  \makebox[0.24\textwidth]{%
      \includegraphics[width=0.12\textwidth]{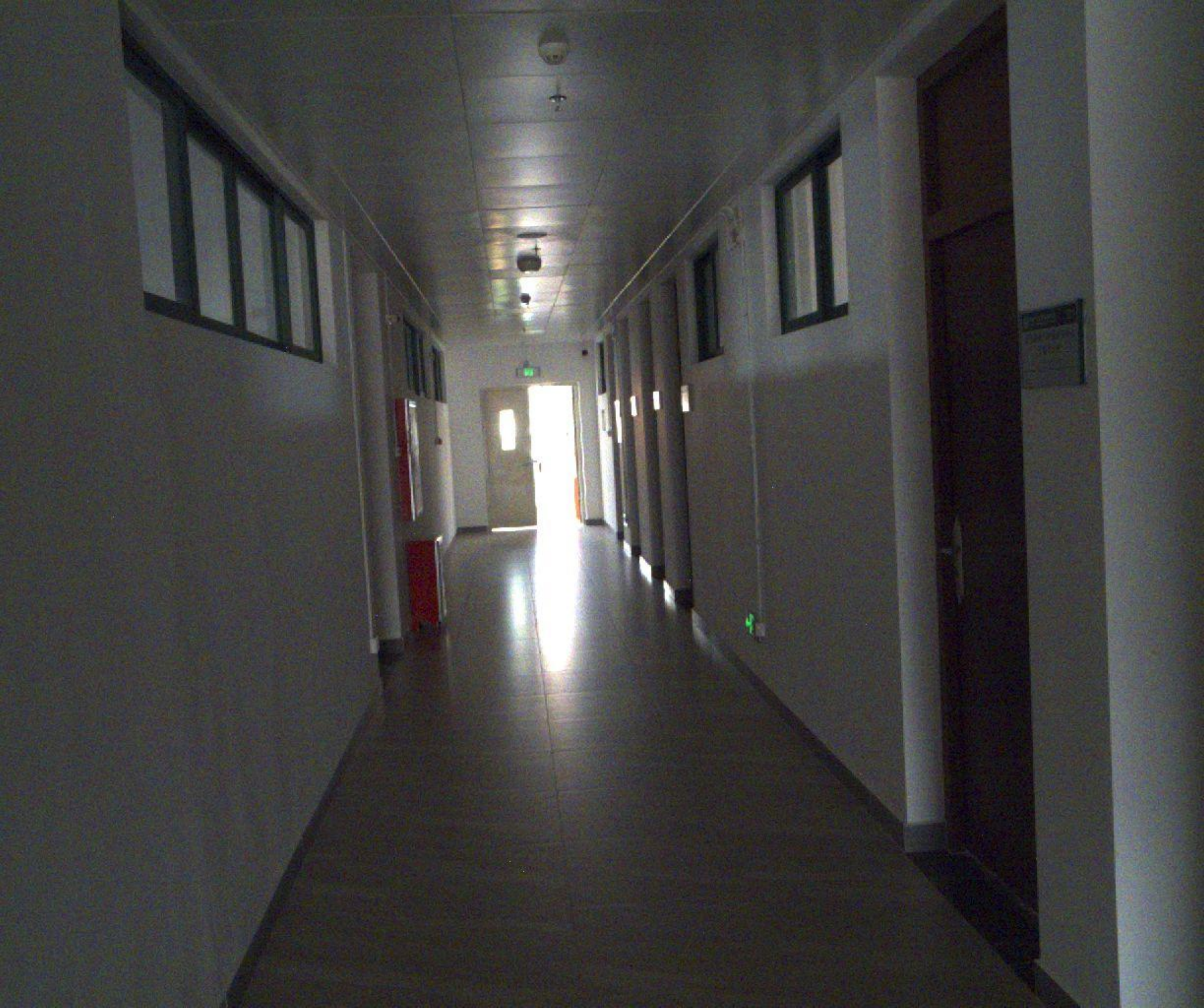}\hspace{\hwidth}
      \includegraphics[width=0.12\textwidth]{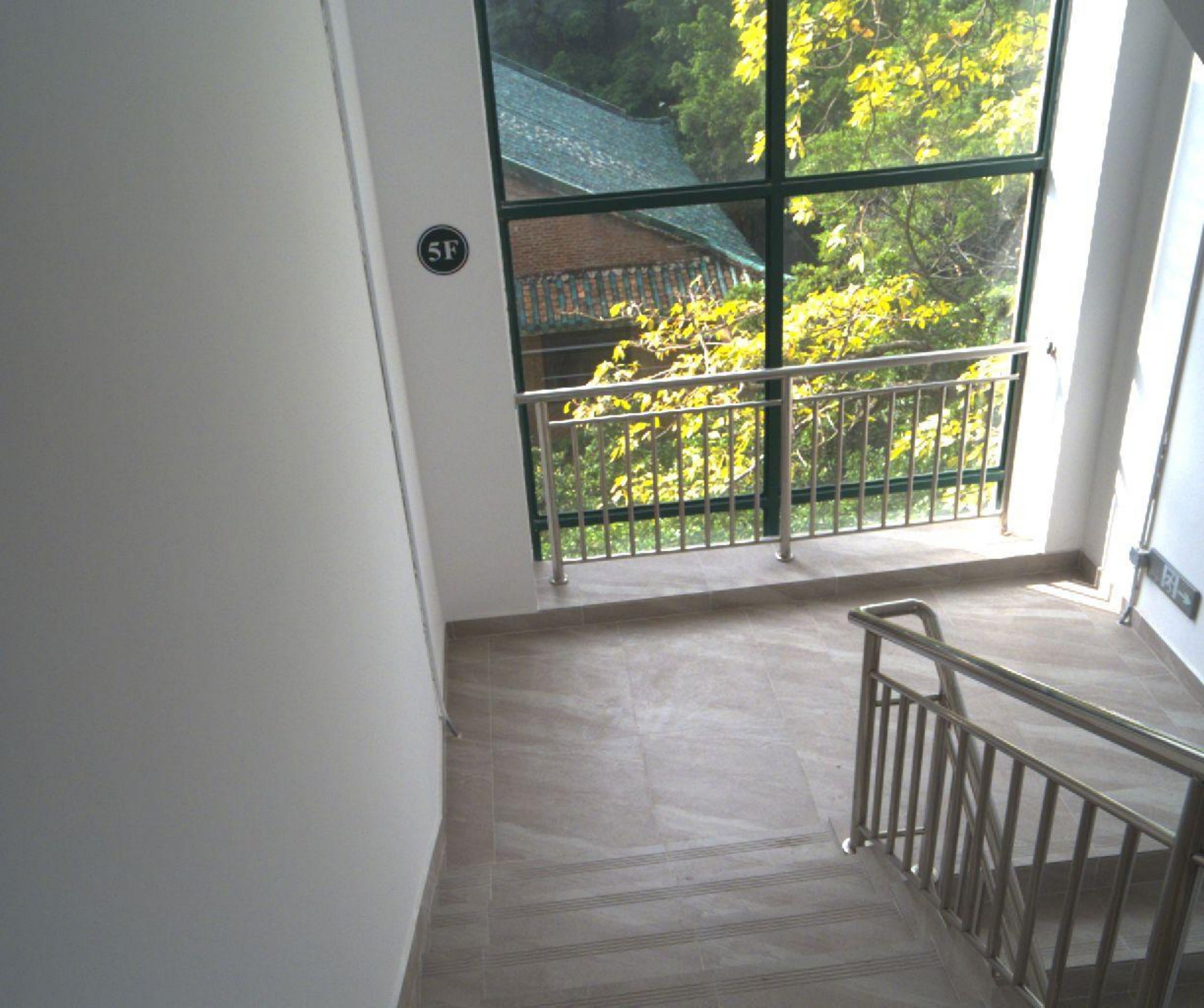}}
  \makebox[0.24\textwidth]{%
      \includegraphics[width=0.12\textwidth]{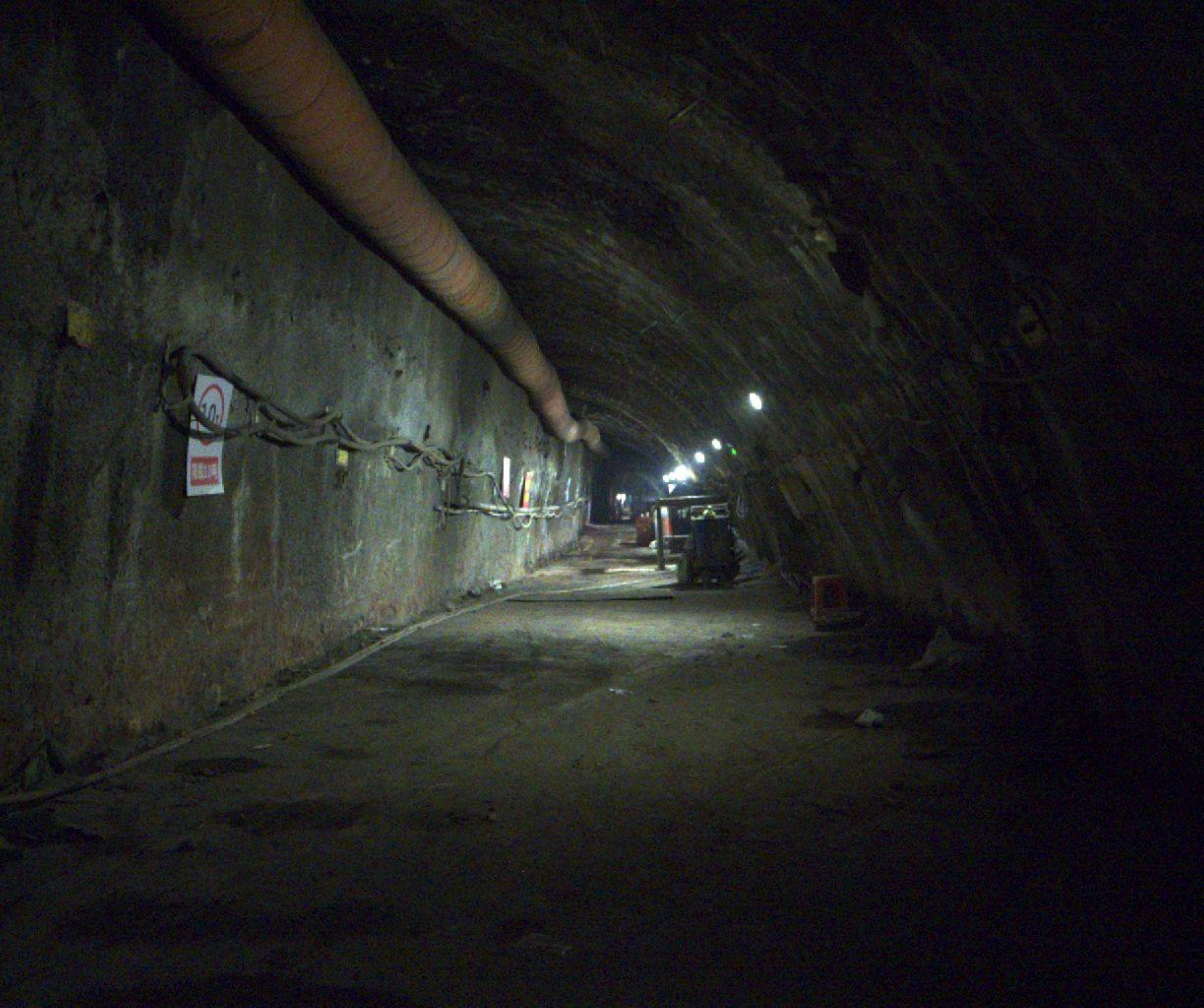}\hspace{\hwidth}
      \includegraphics[width=0.12\textwidth]{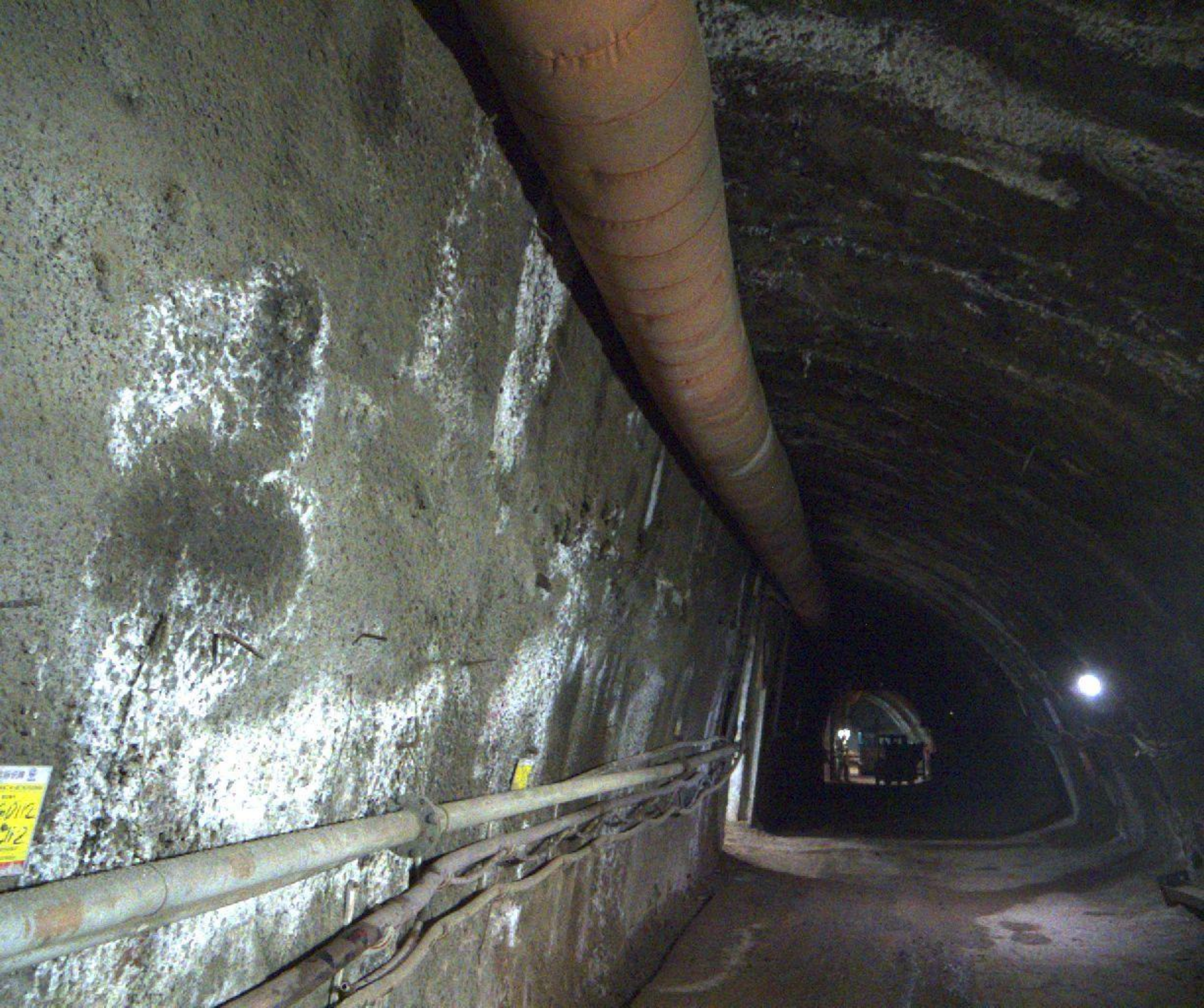}}
  \makebox[0.24\textwidth]{%
      \includegraphics[width=0.12\textwidth]{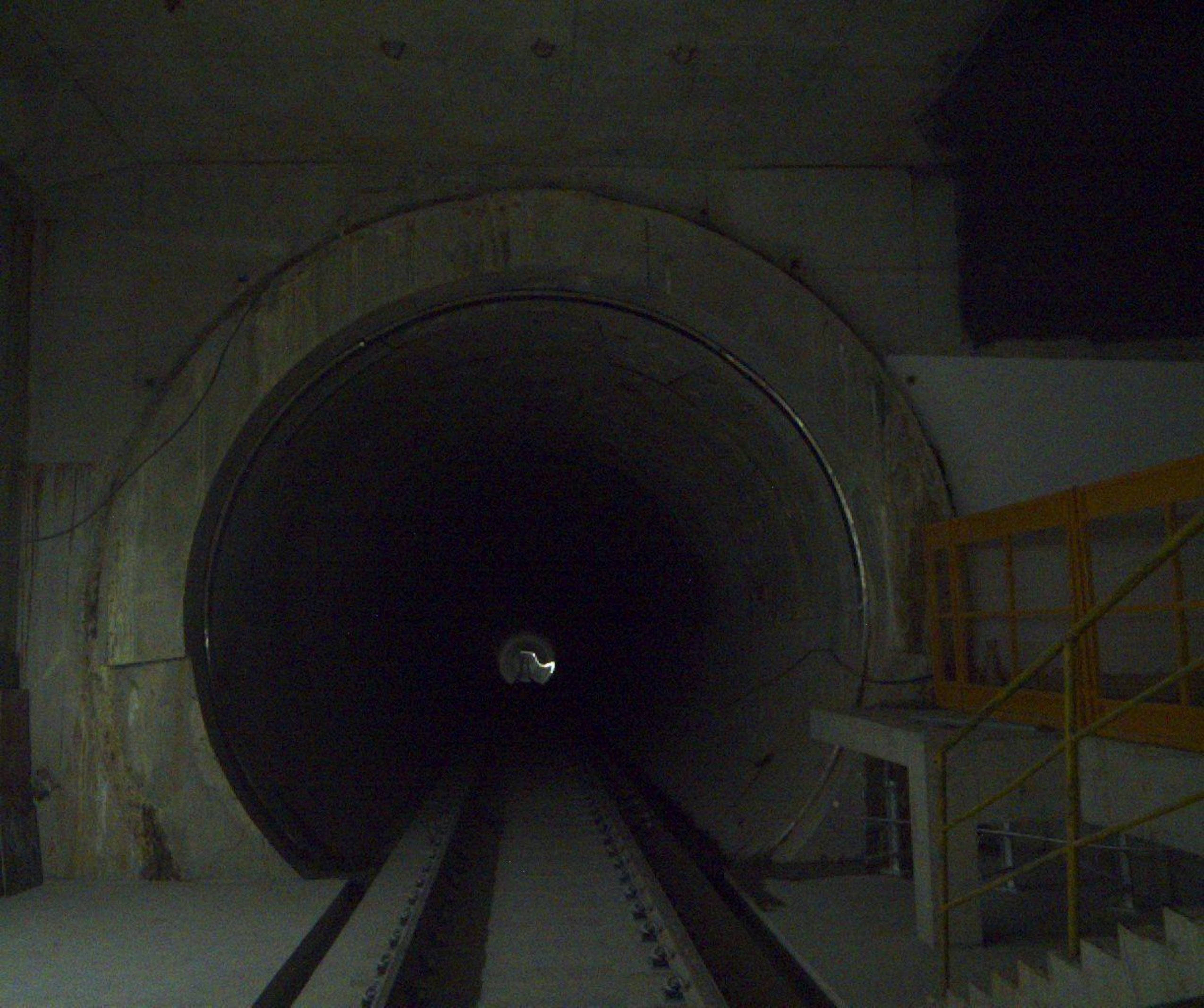}\hspace{\hwidth}
      \includegraphics[width=0.12\textwidth]{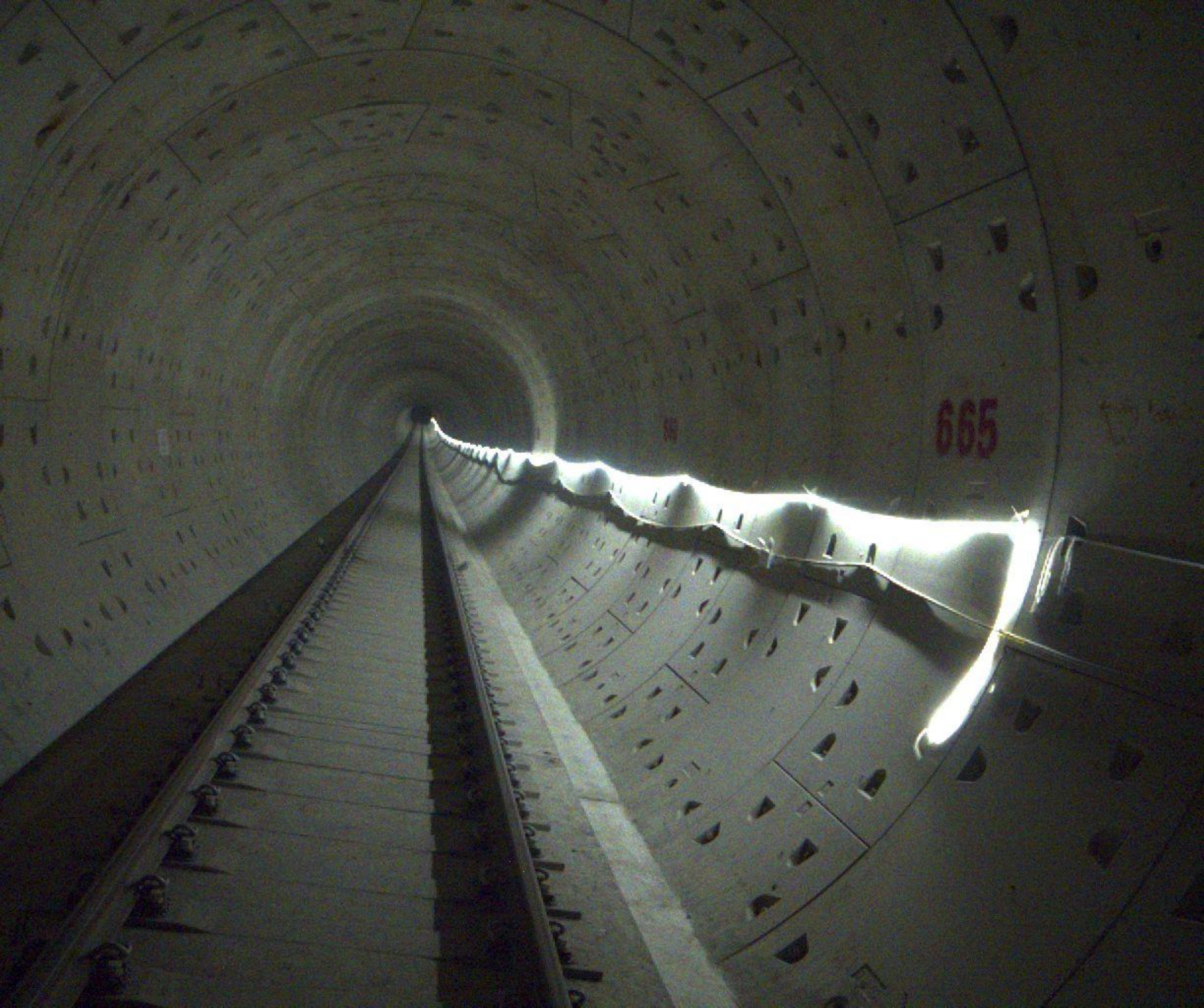}}
  \\
  \makebox[0.24\textwidth]{\small Flat Ground }
  \makebox[0.24\textwidth]{\small Stairs }
  \makebox[0.24\textwidth]{\small Metro Tunnel (Mine tunnelling)} 
  \makebox[0.24\textwidth]{\small Metro Tunnel (Shield method)} 
  \\[2pt]
  \makebox[0.24\textwidth]{%
      \includegraphics[width=0.12\textwidth]{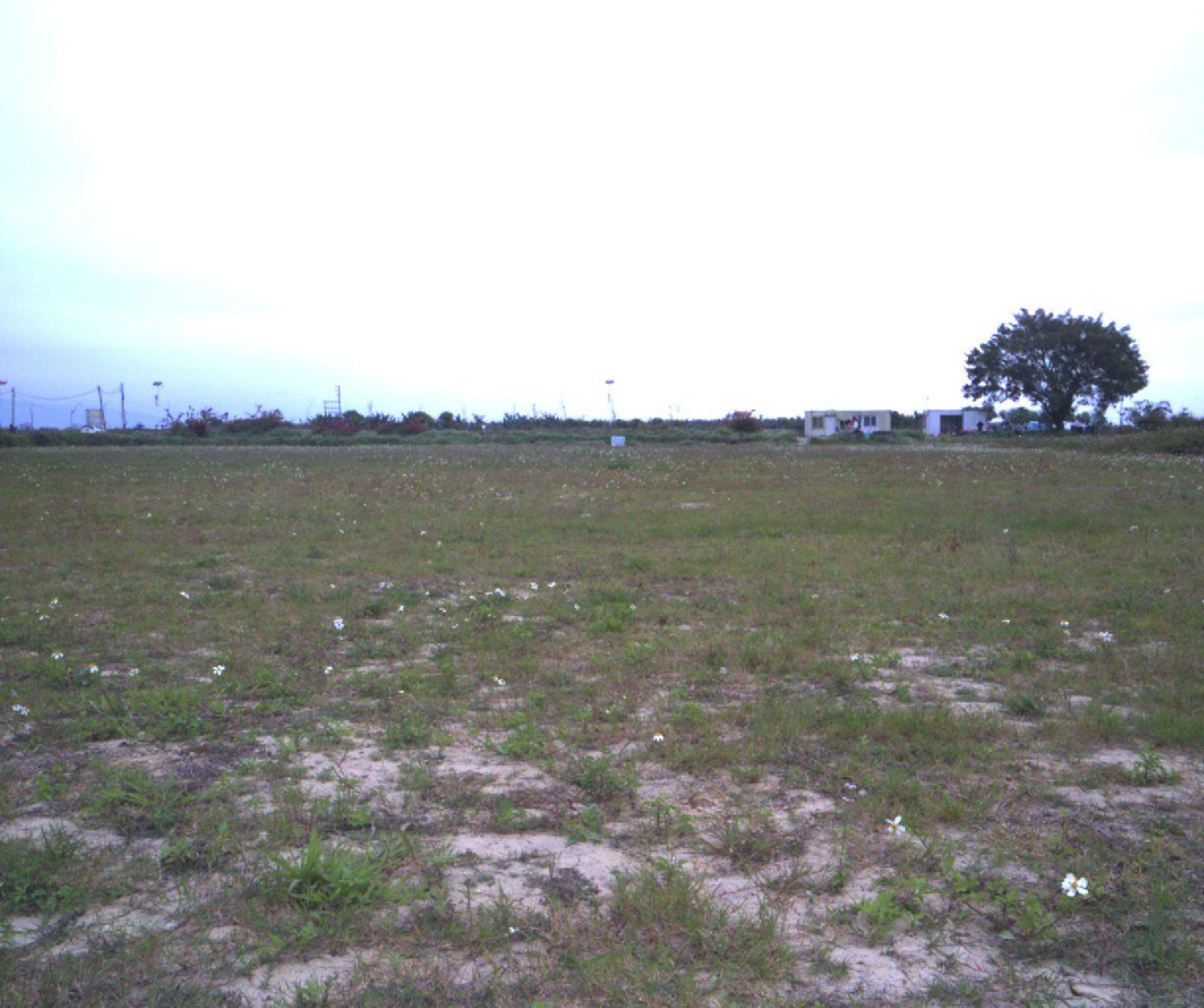}\hspace{\hwidth}
      \includegraphics[width=0.12\textwidth]{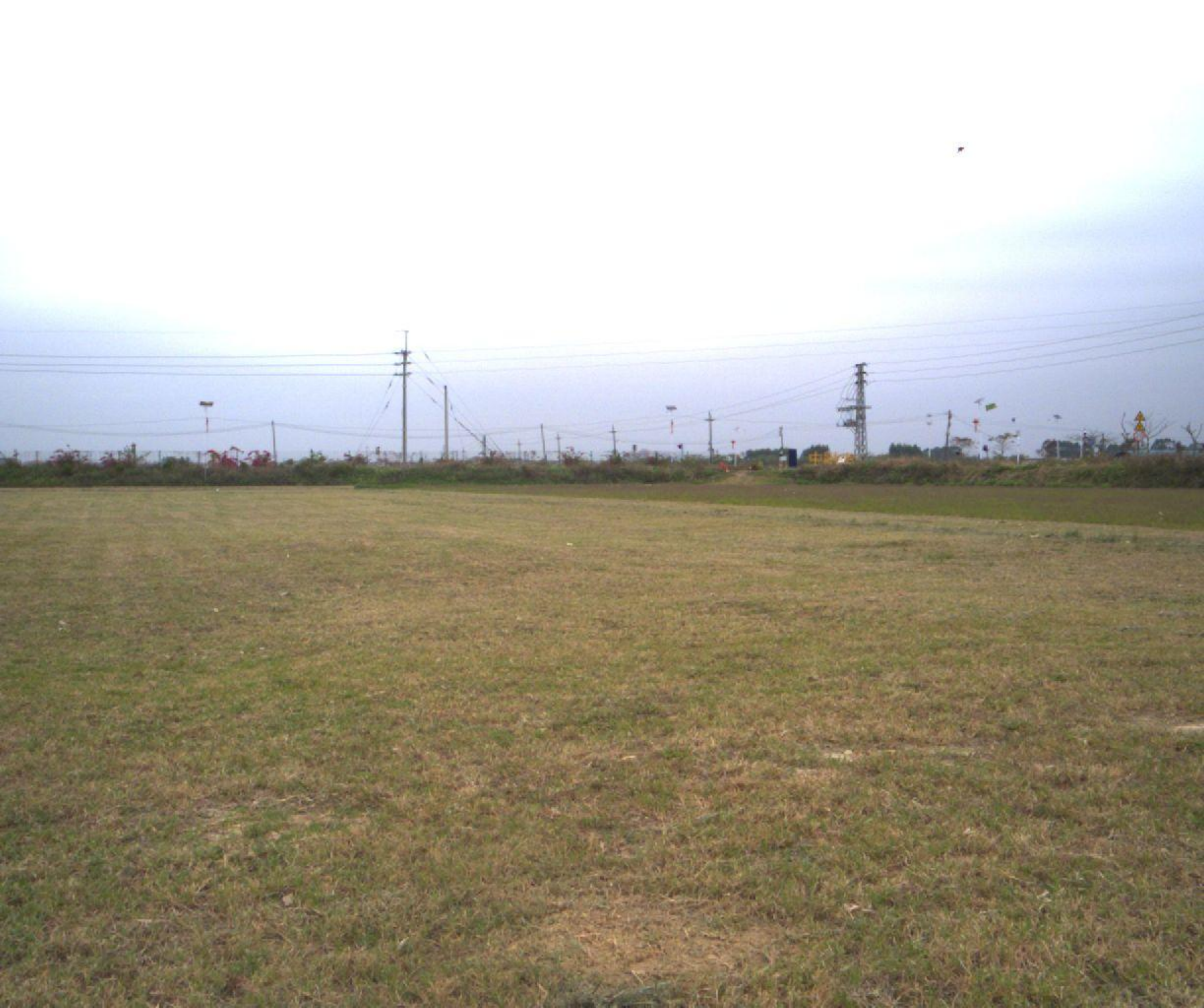}} 
  \makebox[0.24\textwidth]{%
      \includegraphics[width=0.12\textwidth]{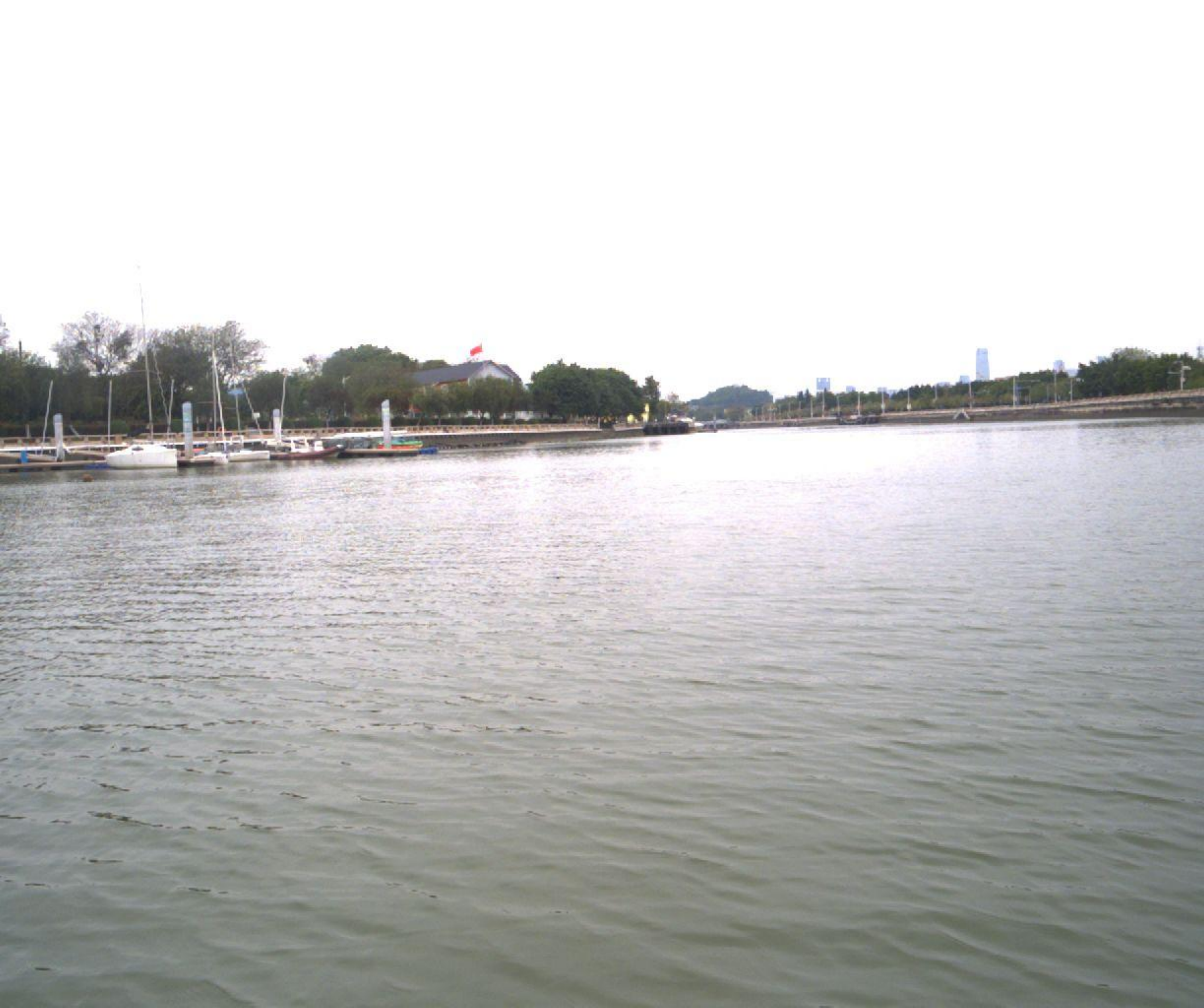}\hspace{\hwidth}
      \includegraphics[width=0.12\textwidth]{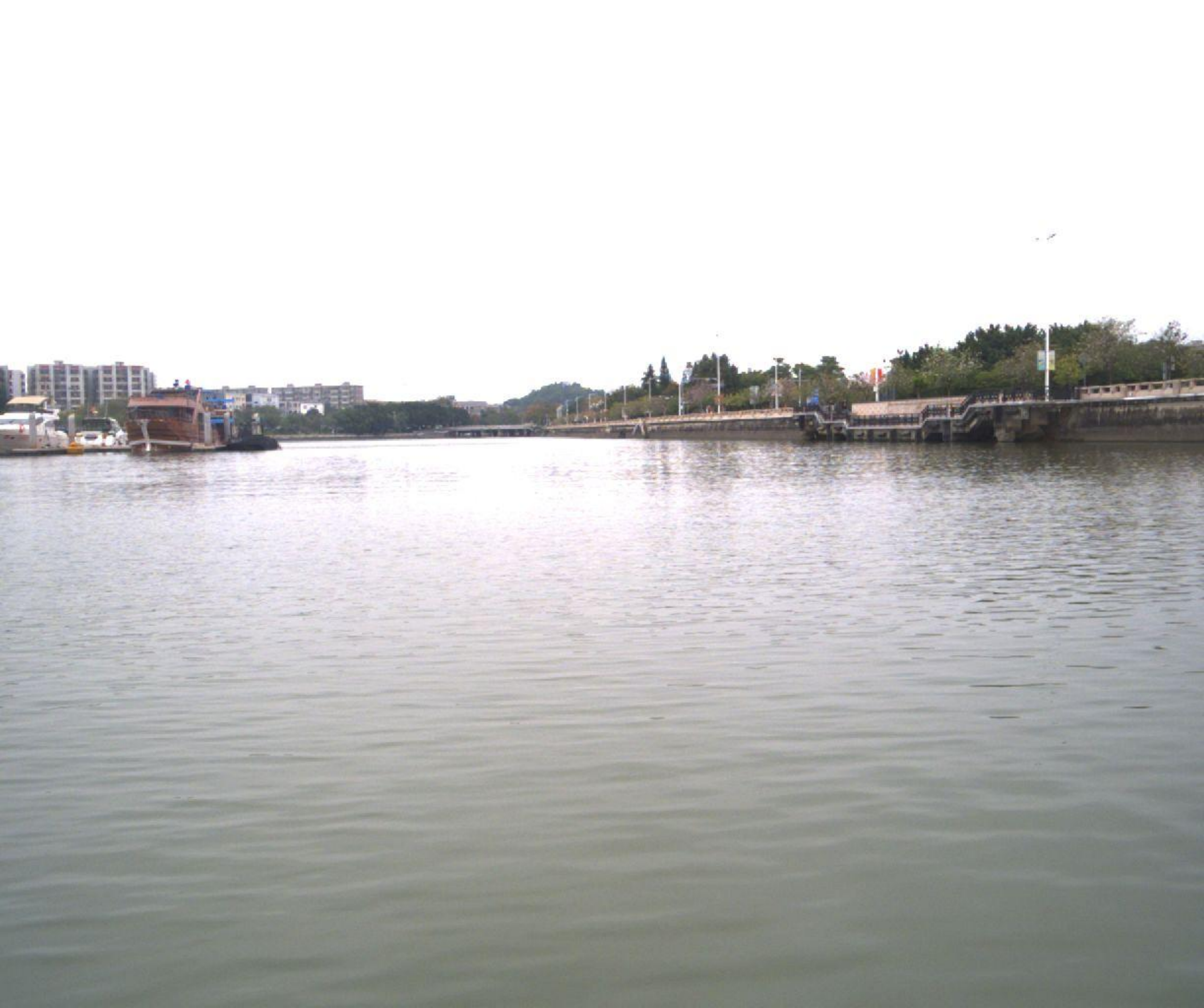}}
  \makebox[0.24\textwidth]{%
      \includegraphics[width=0.12\textwidth]{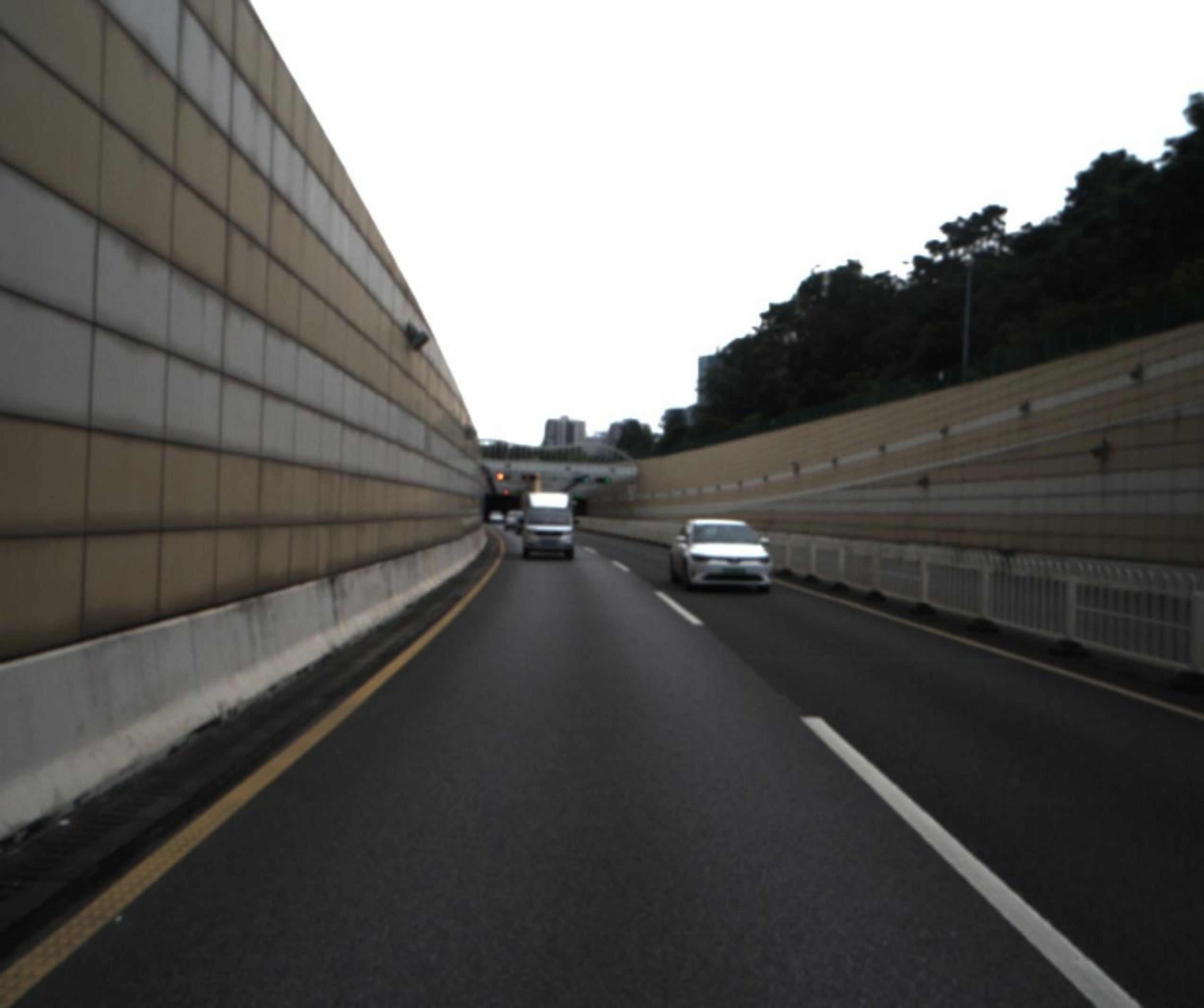}\hspace{\hwidth}
      \includegraphics[width=0.12\textwidth]{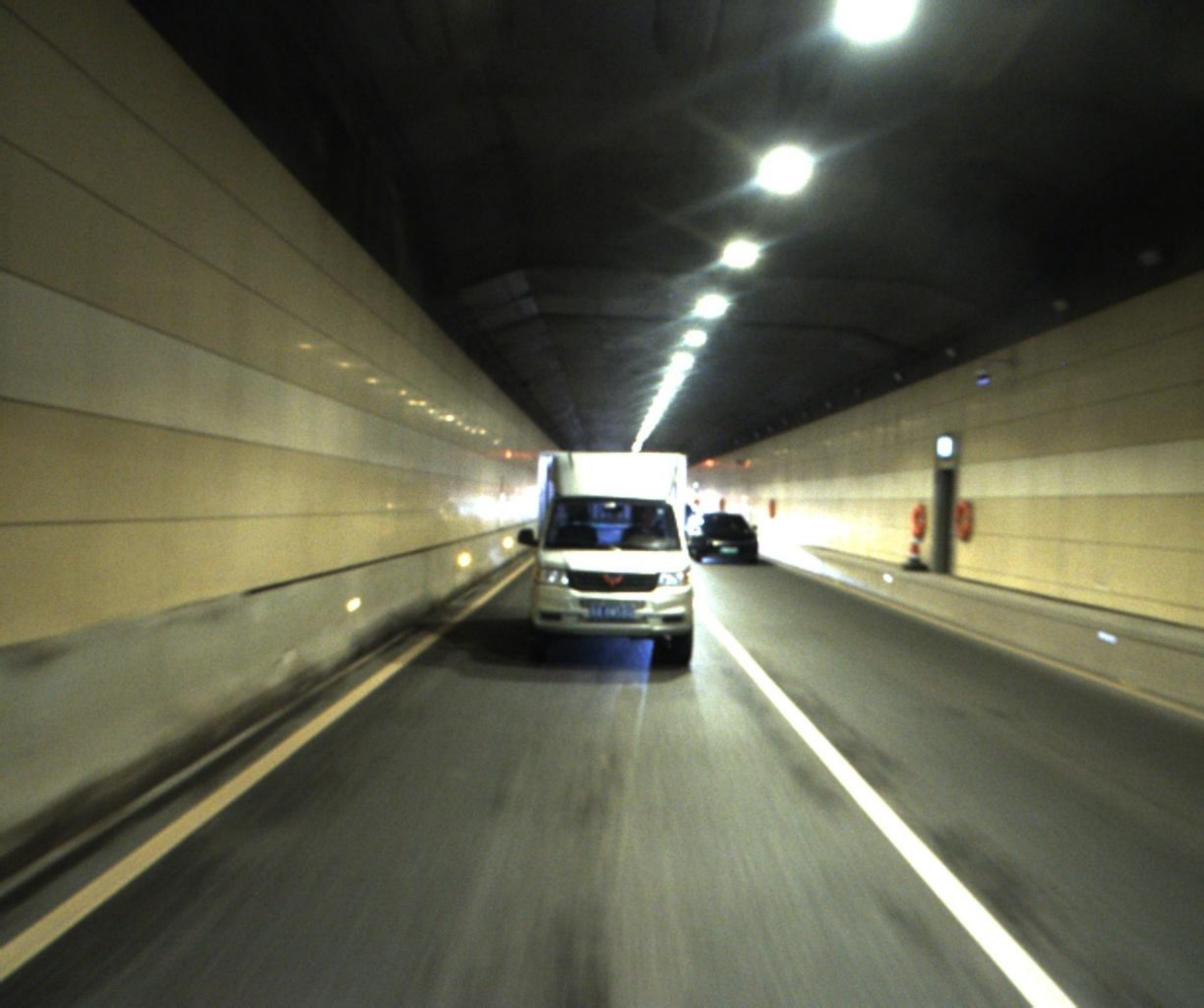}}
  \makebox[0.24\textwidth]{%
      \includegraphics[width=0.12\textwidth]{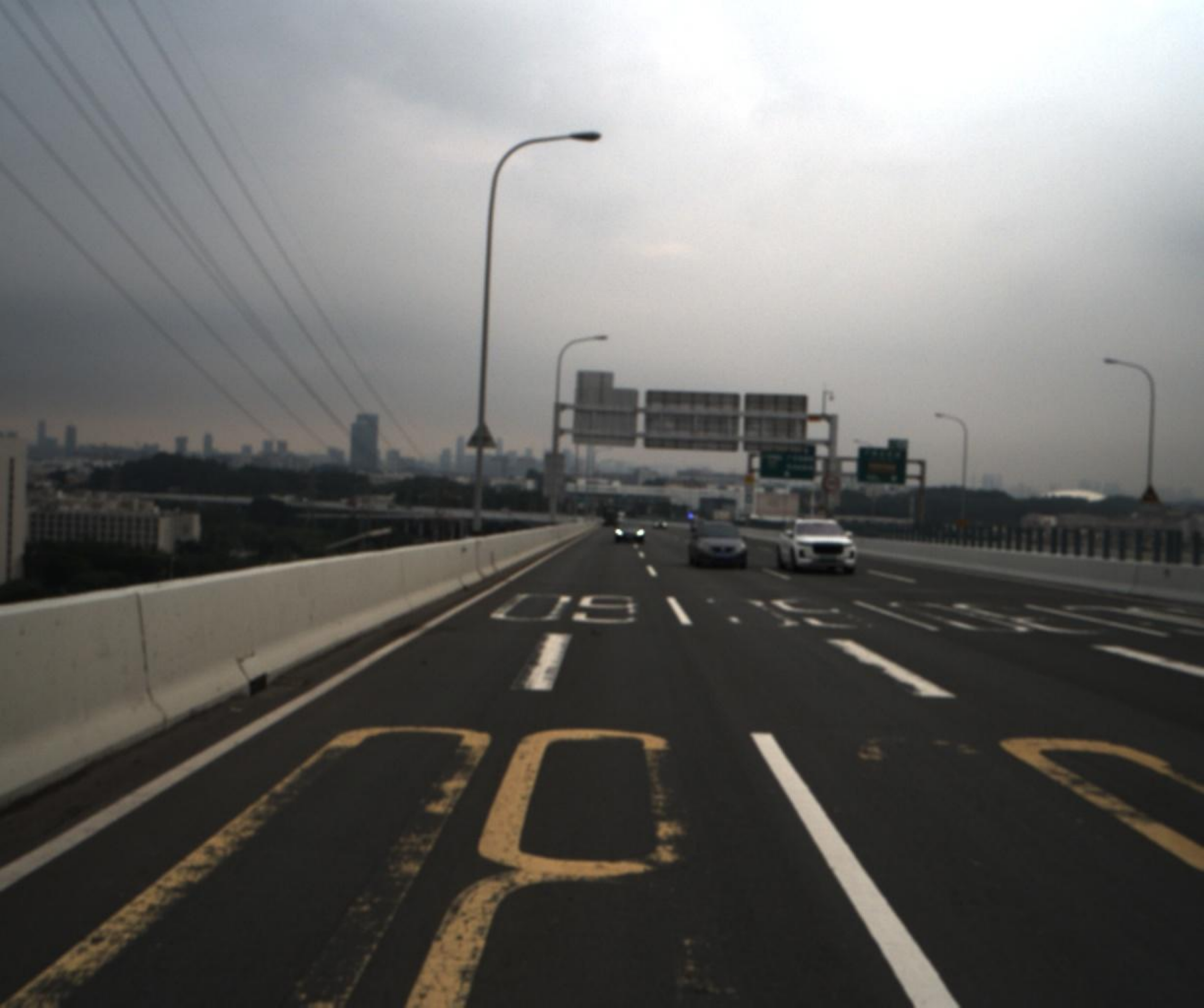}\hspace{\hwidth}
      \includegraphics[width=0.12\textwidth]{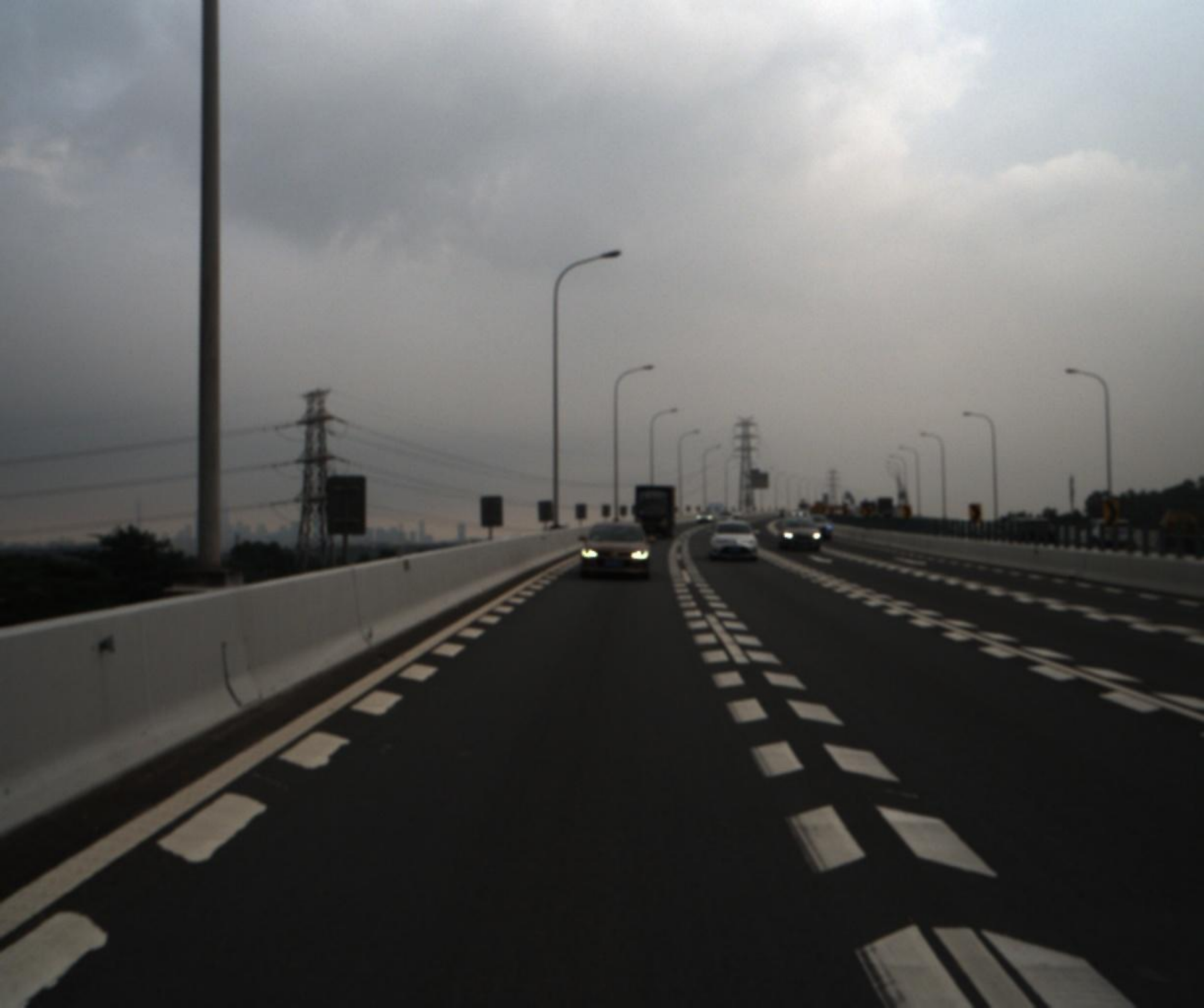}}
  \\
  \makebox[0.24\textwidth]{\small Offroad}
  \makebox[0.24\textwidth]{\small Inland Waterways}
  \makebox[0.24\textwidth]{\small Urban Tunnel} 
  \makebox[0.24\textwidth]{\small Bridges} 
  \caption{Images from different scenes.} 
  \label{fig:Scene_images_in_multiple_sequences}
\end{figure*}

\subsubsection{IMU Calibration:} 
We utilized the imu\_utils \footnote{imu\_utils: \url{https://github.com/gaowenliang/imu_utils}} toolbox to perform calibration on the random walk and Gaussian white noise parameters of the Xsens MTi 30 IMU across the three devices. Furthermore, we employed  LI-Init \citep{Zhu2022RobustRL} for bias calibration of the built-in IMUs in both the Livox and Ouster LiDAR sensors.

\subsubsection{Stereo Cameras Calibration:} 
We captured images of a $12\times9$ checkerboard using the sensor suite. The camera intrinsics and stereo extrinsic parameters were calibrated using a Matlab toolbox, employing a pinhole camera model and a radial-tangential distortion model. The calibration results were obtained by minimizing the reprojection error and eliminating outliers with significant errors.

\subsubsection{Camera-IMU Extrinsic Calibration:} Utilizing the intrinsics of cameras and IMU, we employ the Kalibr \footnote{Kalibr: \url{https://github.com/ethz-asl/kalibr}} toolbox to perform spatial calibration of the IMU concerning all cameras. The process necessitated capturing the sensor suite's 6 degrees of freedom motion to ensure accuracy in the calibration results.

\subsubsection{Camera-LiDAR Extrinsic Calibration:} We employ Matlab's lidarCameraCalibrator to calibrate Extrinsic between the mechanical LiDAR and the camera for devices $\alpha$ and $\beta$. For device $\gamma$, we utilize joint-lidar-camera-calib \citep{Li2023JointIA} to calibrate Livox Avia and the camera without checkerboards.

\subsubsection{LiDAR-IMU Extrinsic Calibration:} 
We performed external calibration of the LiDAR and Xsens MTi 30 IMU for the three devices using Li-Init \citep{Zhu2022RobustRL}. Additionally, we conducted external parametric calibration for the Livox Avia and Ouster sensors along with their built-in IMUs.

%% file: chapter/dataset_description.tex
\section{DATASET}
\label{sec:dataset_description}
\runninghead{CHEN \textit{et~al.}}

The dataset encompasses a diverse range of geometrically degenerate scenarios, including flat surfaces, stairs, subway tunnels, off-road terrain, inland waterways, urban tunnels, and bridges. These scenarios present different levels and types of geometric degeneration, both translational and rotational. It includes data from various LiDAR sensors and platforms operating in challenging environments. The dataset consists of 64 sequences designed for algorithm development and evaluation, summarized in Table \ref{tab:overview_of_dataset}. These sequences feature both unidirectional forward trajectories and looping trajectories. Additional details about the trajectories can be found on the dataset’s homepage. Figure \ref{fig:Scene_images_in_multiple_sequences} offers a visual overview of the scenarios.

\begin{figure*}[t]
    \centering
    \includegraphics[width=0.99\textwidth]{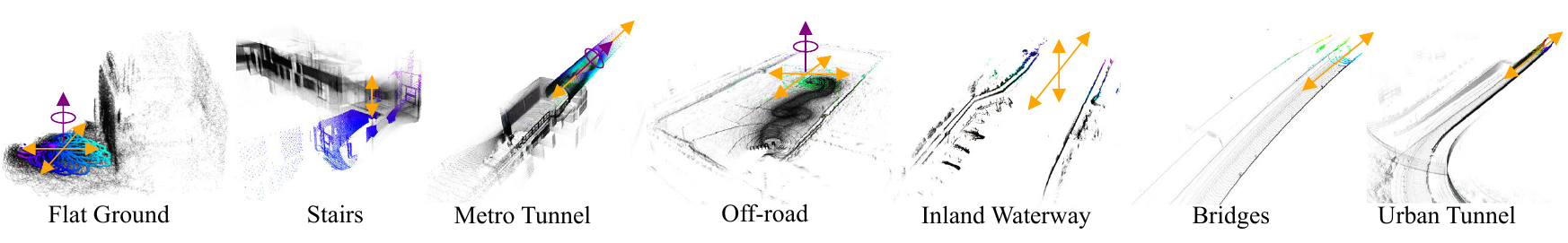}
    \caption{Visualization of multi-degenerate scenarios in LiDAR point cloud data: representation of spatial map and current frame degradations. The translational degradation is denoted by \textcolor{orange}{the orange arrow $\leftrightarrow$}, while the rotational degradation is signified by \textcolor{purple}{the purple arrow $\rightarrow $}.
    }
    \label{fig:geode_scenario}
  \end{figure*}

\begin{figure*}[t]
    \centering
    \includegraphics[width=0.99\textwidth]{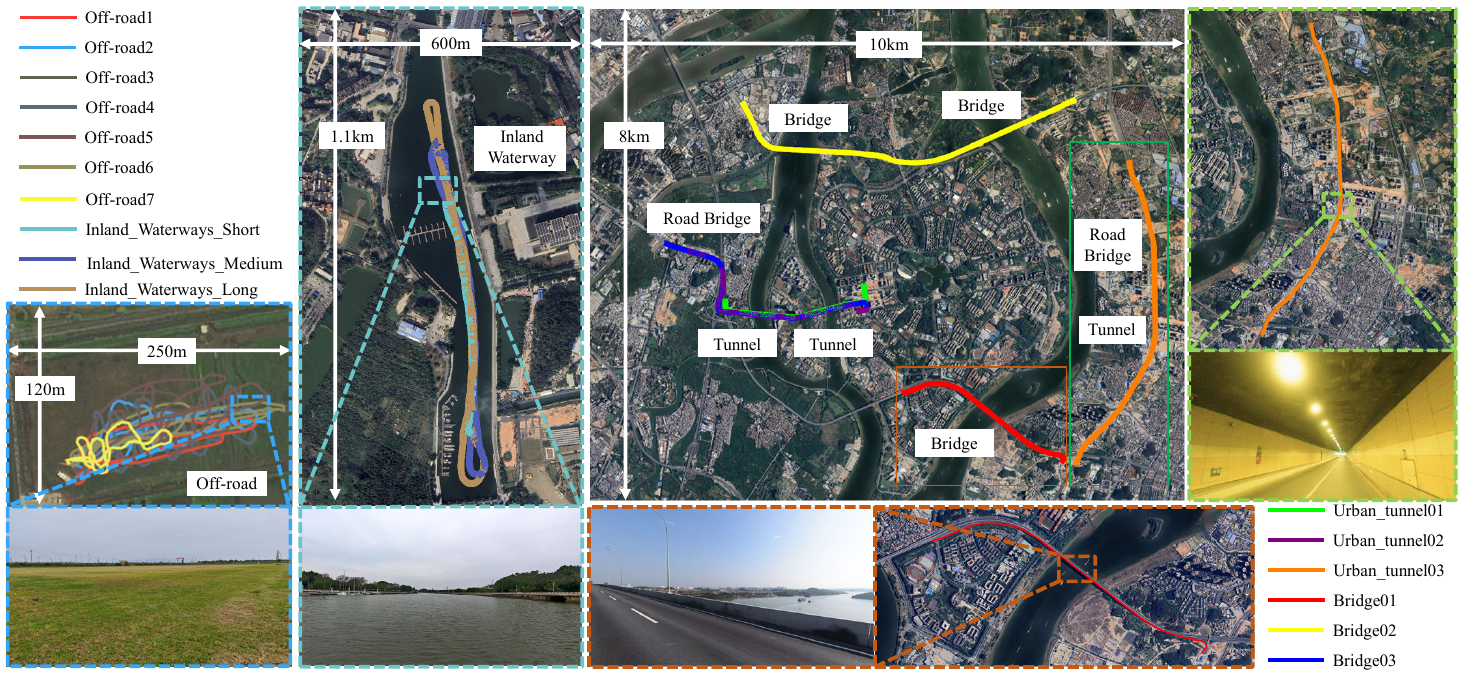}
    \caption{Trajectories of multiple sequences captured in diverse outdoor scenarios, encompassing environments with varying scales and degrees of degradation, such as off-road areas, inland waterways, urban tunnels, and bridges.}
    \label{fig:Urban_Tunnel_Sequences_visualization}
\end{figure*}

\begin{figure*}[t]
    \centering
    \subfloat[\texttt{Handheld\_Flat\_surfaces\_aggressive}]{
        \label{fig:handheld_motion}
        \centering
        \includegraphics[width=0.99\linewidth]{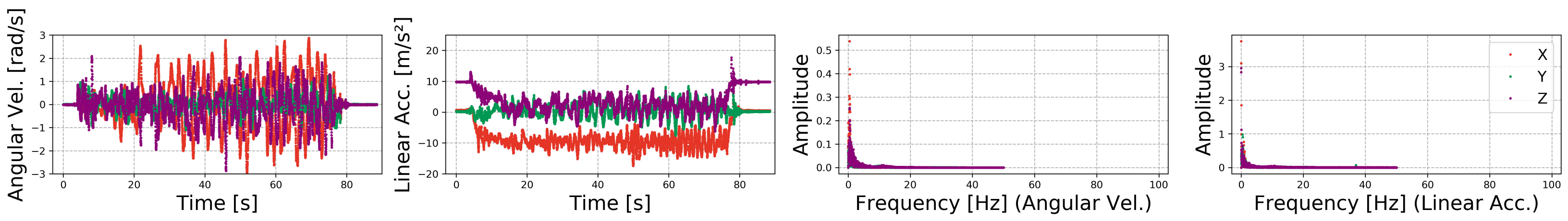}
    }

    \subfloat[\texttt{Sailboat\_Inland\_Waterways\_Medium\_Alpha}]{
        \label{fig:legged_motion}
        \centering
        \includegraphics[width=0.99\linewidth]{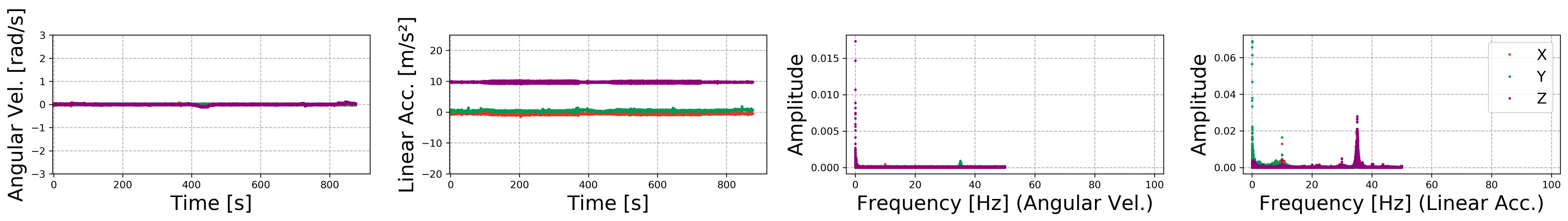}
    }

    \subfloat[\texttt{Ugv\_Offroad1\_alpha\_rect}]{
        \label{fig:ugv_motion}
        \centering
        \includegraphics[width=0.99\linewidth]{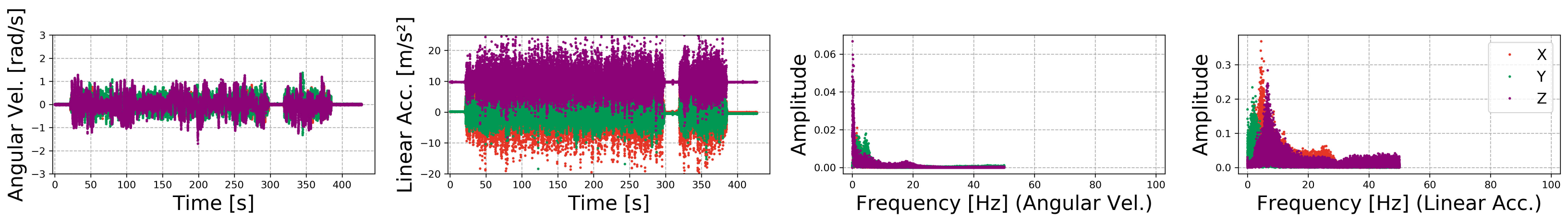}
    }

    \subfloat[\texttt{Vehicle\_Urban\_Tunnel01}]{
        \label{fig:vehicle_motion}
        \centering
        \includegraphics[width=0.99\linewidth]{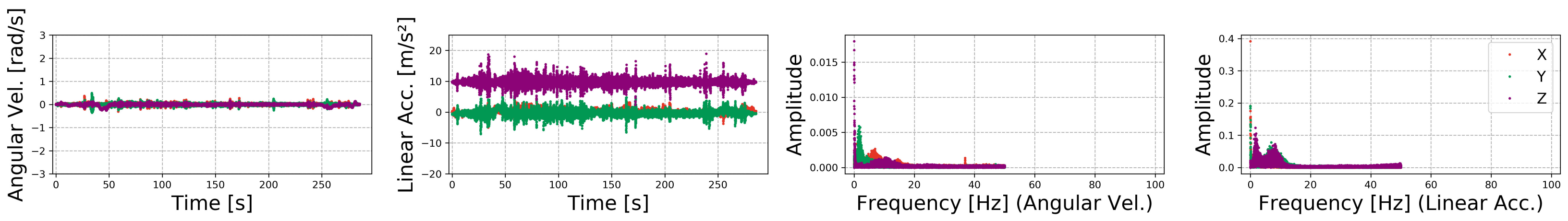}
    }
    \caption{Cross-platform dynamic behavior and motion feature analysis in the GEODE dataset: insights from IMU data visualization. The motion characteristics are explored by leveraging visualization tools within the FusionPortableV2 Dataset, as referenced in \citep{Wei2024FusionPortableV2AU}. 
    }
    \label{fig:platforms_motion_pattern}
\end{figure*}

\subsection{Scenarios and Challenges}
\subsubsection{Flat Surfaces:} 
Non-repetitive LiDARs with limited FoV face challenges when perceiving a single plane, such as walls or the ground. This leads to geometric degeneration in two translational dimensions (along the plane) and one rotational dimension (across the plane). In this scenario, we employed the Livox Avia LiDAR facing the ground, capturing sequences with both slow and aggressive motions.

\subsubsection{Stairs:} Stairs are confined spaces characterized by parallel planes (steps) and perpendicular planes (walls). The limited vertical FoV of sensors diminishes the information content, making scans less distinguishable. We utilized three devices with distinct scan lines and varying FoVs to gather data from corridors and stairs across multiple floors. Each device followed a path descending one of the stairs from the seventh floor and returning via the other one, creating a loop.

\subsubsection{Metro Tunnels:} Subway tunnels exhibit geometric degeneration in both translational and rotational dimensions along the axial direction. Data collection within these tunnels utilized two construction methods: the shield method and the mine tunneling method. With the shield method, tunnel walls exhibited smooth surfaces, representing a highly degenerate scenario. In contrast, the mine tunneling method introduced irregular wall shapes, reducing the degree of geometric degeneration. A loop was formed by traversing several hundred meters along the tunnel and then doubling back. Some sequences were collected using multiple LiDAR sensors simultaneously to capture more spatial features.

\subsubsection{Inland Waterways:} Water surface reflections challenge LiDAR SLAM implementations, potentially disrupting the LiDAR signal and causing noisy or inaccurate measurements. Additionally, open water environments may have fewer laser points or sparsely distributed features. Water absorption along the z-axis weakly constrains the point cloud, potentially leading to drift phenomena and slight performance degradation during movement along this axis. Sequences include one-way forward, forward and back, and forward and back again paths. All heterogeneous LiDAR data were collected simultaneously in this scenario.

\subsubsection{Offroad:} A UGV was utilized as a data collection platform on an extensive grass track, with three devices deployed to capture data during both slow and vigorous movements. The offroad scenario demonstrates similar geometric degeneration to flat surfaces; however, long-distance traversal imposes greater demands on the algorithm to overcome such degeneration. The UGV completed a loop by returning to the starting point after collecting respective trajectory. All heterogeneous LiDAR data were collected simultaneously in this scenario.

\subsubsection{Urban Tunnel:} Scan matching in urban tunnels is challenging due to the absence of distinct features or landmarks. These environments exhibit geometric degeneration, exacerbated by the presence of highly dynamic moving vehicles, significantly affecting point cloud registration. Furthermore, the low-texture surroundings within the tunnel and abrupt light variations during entry and exit impede camera effectiveness in compensating for LiDAR degeneracy.

\subsubsection{Bridge:} The bridge is flanked solely by the river on both sides, lacking distinctive landmarks for localization. Bridge scenes display limited textural variation, and the feature data in these areas tends to be uniform and repetitive, contributing to the challenge of achieving precise positioning.

\subsection{Key Features}

\subsubsection{Heterogeneous LiDAR:}
The GEODE dataset includes data from both non-repetitive and spinning LiDARs, each with varying fields of view and scan lines. Some of these LiDARs provide additional channels of information beyond geometric data, enhancing localization in geometrically degraded environments.
In various applications, different types of LiDAR systems may be required to meet specific needs. For example, autonomous vehicles may necessitate high-resolution mechanical LiDAR for precise environmental perception, whereas lightweight robots might be better suited with non-repetitive LiDAR due to their reduced weight. A dataset that includes a variety of LiDAR data can assist developers in selecting the appropriate SLAM algorithms for specific hardware configurations or in customizing algorithms to adapt to the unique characteristics of each hardware setup.  
This diversity ensures comprehensive testing across various LiDAR technologies, improving localization robustness in different hardware environments, especially in geometrically degenerate scenarios.

\subsubsection{Multi-Degenerate Scenarios:}
Geometric degradation in SLAM can be categorized into three types: rotational, translational, and a combination of both. As depicted in Figure \ref{fig:geode_scenario}, the GEODE dataset provides a rich tapestry of these degraded contexts, offering a diverse set of challenges for SLAM algorithms to overcome.
The importance of a multi-degenerate dataset is further underscored by the deployment of algorithms across a diverse array of robotic systems engaged in various tasks, such as autonomous navigation and swarm-based search operations. These tasks often encounter a range of unpredictable degradation conditions. A dataset enriched with a wide variety of these degradation scenarios, as provided by GEODE, is crucial for equipping algorithms to better anticipate and adapt to the challenges encountered in practical applications.
Moreover, the adaptability of SLAM algorithms is enhanced by the inclusion of environments spanning various scales. The GEODE dataset, ranging from indoor rooms to highways, ensures that the algorithms are not only tested in confined spaces but are also prepared for the complexities of large-scale outdoor deployments. This comprehensive approach is essential for developing robust SLAM solutions that maintain efficacy and precision across different operational contexts.

Furthermore, the outdoor sequences of the GEODE dataset, as presented from a satellite perspective in Figure \ref{fig:Urban_Tunnel_Sequences_visualization}, showcase the dataset's extensive coverage, thereby reinforcing the robustness of SLAM algorithms in handling real-world scenarios. The ability to manage such a breadth of environments and degradation types is a testament to the importance of multi-degenerate scenarios in the evolution of SLAM technology.

\subsubsection{Diverse Platform Characteristics:}
The GEODE dataset was collected across four unique platforms: a handheld device, an UGV, a sailboat, and a vehicle. Each platform exhibits unique motion characteristics, including speed, angular velocity, and dynamic frequency, as well as varying operational ranges. As depicted in Figure \ref{fig:platforms_motion_pattern}, the time-domain data captures the transient dynamics of each platform, while the frequency-domain data accentuates the salient features of their motion signatures. The handheld device is characterized by swift fluctuations in velocity and directional shifts. The UGV demonstrates motion profiles marked by high-frequency oscillations and abrupt jolts, attributable to the undulating terrain it navigates. In contrast, the sailboat's motion is governed by the ebb and flow of water currents, resulting in a syncopated mix of low-frequency drifts interspersed with high-frequency oscillations. The vehicular platform contributes data indicative of extensive, rapid transits, occasionally punctuated by sudden variations in pace and trajectory due to the exigencies of traffic and road conditions. This diverse dataset necessitates the development of robust SLAM algorithms capable of accommodating the varied motion profiles. The algorithms must be sensitive to the quick, erratic movements of handheld devices, the stable progression of UGVs, the oscillating patterns of sailboats, and the consistent velocities of vehicles. The GEODE dataset's comprehensive capture of these dynamics is crucial for creating SLAM solutions that are adaptable and reliable in a wide range of real-world scenarios.

\newcommand{\foldericon}{
  \raisebox{-0.2\width}{\includegraphics[height=0.5cm]{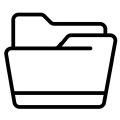}}
}
\newcommand{\txticon}{
  \raisebox{-0.1\width}{\includegraphics[height=0.5cm]{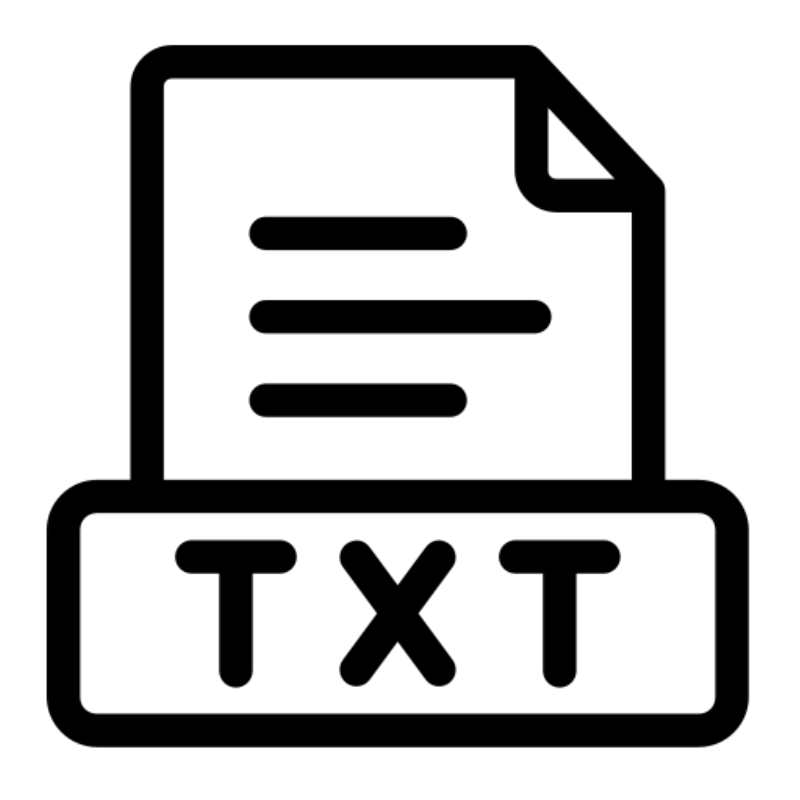}}
}

\newcommand{\yamlicon}{
  \raisebox{-0.1\width}{\includegraphics[height=0.5cm]{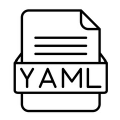}}
}

\newcommand{\binicon}{
  \raisebox{-0.1\width}{\includegraphics[height=0.5cm]{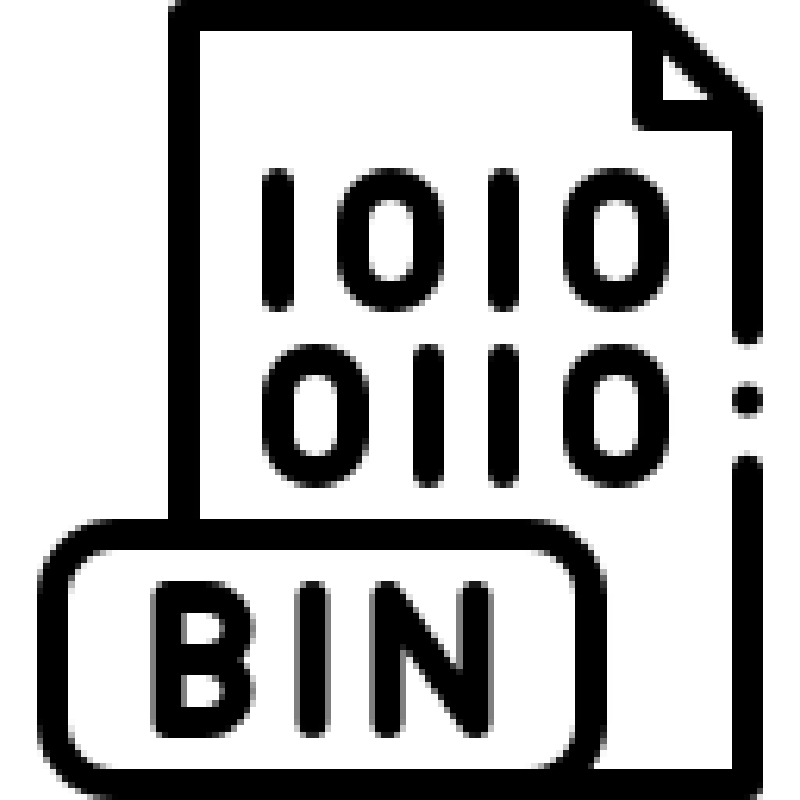}}
}

\newcommand{\dataseticon}{
  \raisebox{-0.4\width}{\includegraphics[height=0.6cm]{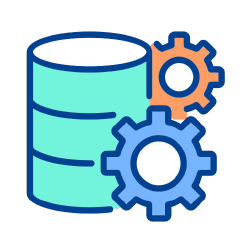}}
}

\newcommand{\cloundicon}{
  \raisebox{-0.2\width}{\includegraphics[height=0.5cm]{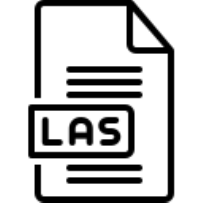}}
}

\newcommand{\rosicon}{
  \raisebox{-0.1\width}{\includegraphics[height=0.35cm]{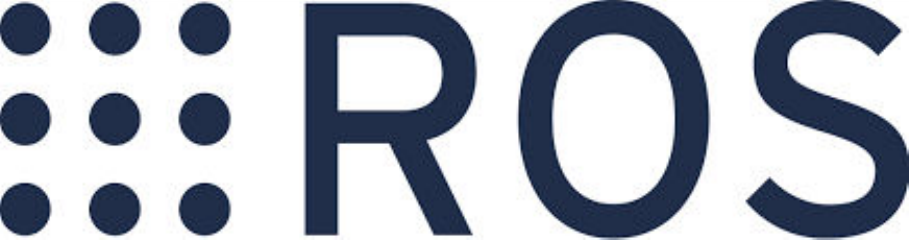}}
}

\newcommand{\zipicon}{
  \raisebox{-0.2\width}{\includegraphics[height=0.5cm]{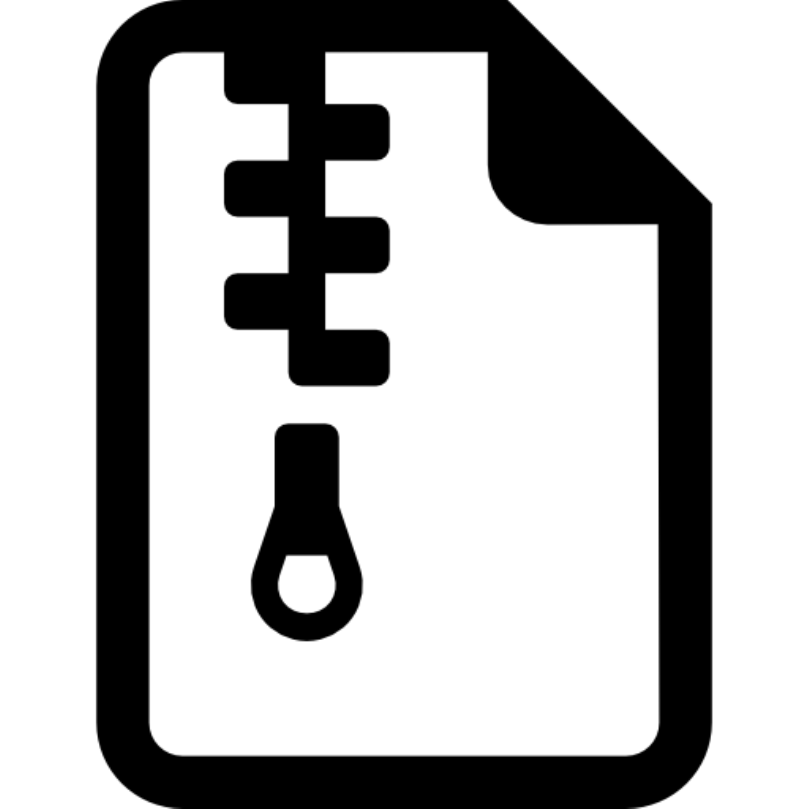}}
}

\newcommand{\imgicon}{
  \raisebox{-0.2\width}{\includegraphics[height=0.5cm]{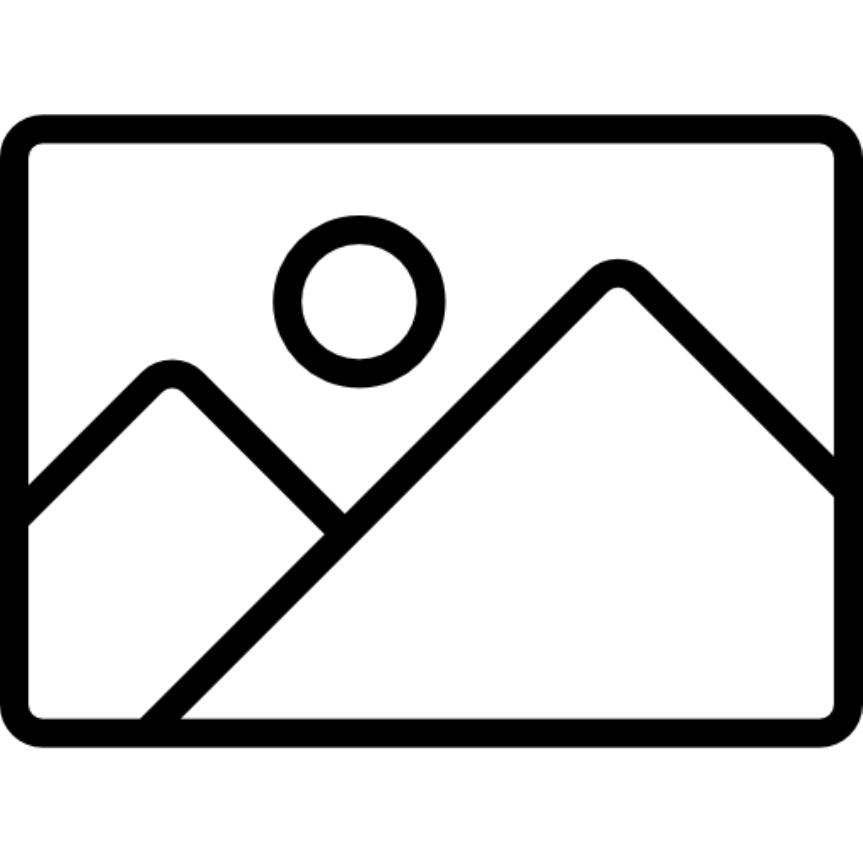}}
}
\begin{figure}[t]
  \dirtree{%
  .1 \hspace{-0.6cm} \dataseticon GEODE/.
  .2 \hspace{-0.6cm} \foldericon sensor data.
  .3 \hspace{-0.6cm} \foldericon <map\_env>/<sequence\_id>.
  .4 \hspace{-0.3cm} \rosicon <sequence\_id>.bag.
  .4 \hspace{-0.4cm} \zipicon <sequence\_id>.zip.
  .4 \hspace{-0.6cm} \foldericon LiDAR.
  .5 \hspace{-0.6cm} \foldericon bin.
  .6 \hspace{-0.4cm} \binicon <timestamp>.bin.
  .5 \hspace{-0.6cm} \foldericon depth\_image$^{\bm{*}}$.
  .6 \hspace{-0.4cm} \imgicon <timestamp>.jpg.
  .5 \hspace{-0.6cm} \foldericon reflectivity\_image$^{\bm{*}}$.
  .6 \hspace{-0.4cm} \imgicon <timestamp>.jpg.
  .5 \hspace{-0.6cm} \foldericon IMU$^{\diamondsuit}$.
  .6 \hspace{-0.4cm} \txticon imu.txt.
  .4 \hspace{-0.6cm} \foldericon Camera.
  .5 \hspace{-0.6cm} \foldericon image\_left.
  .6 \hspace{-0.4cm} \imgicon <timestamp>.jpg.
  .5 \hspace{-0.6cm} \foldericon image\_right.
  .6 \hspace{-0.4cm} \imgicon <timestamp>.jpg.
  .4 \hspace{-0.6cm} \foldericon IMU.
  .5 \hspace{-0.4cm} \txticon imu.txt.
  .2 \hspace{-0.6cm} \foldericon calibration\_files/.
  .3 \hspace{-0.4cm} \yamlicon <device\_id>.yaml.
  .2 \hspace{-0.6cm} \foldericon groundtruth.
  .3 \hspace{-0.6cm} \foldericon map.
  .4 \hspace{-0.4cm} \cloundicon <map\_env>.las.
  .3 \hspace{-0.6cm} \foldericon traj.
  .4 \hspace{-0.4cm} \txticon <sequence\_id>.txt.
  } 
  \caption[xx]{File structure of the GEODE datase.$^{\bm{*}}$Depth and reflection images are exclusively accessible within the dataset when captured by the $\beta$ device. Additionally, the built-in IMU data, denoted by $^{\diamondsuit}$, is specifically available for the Ouster and Livox LiDAR.}
  \label{fig:dataset_format}
\end{figure}

\subsection{Dataset Organization}
\label{sec:dataset_organization}

Figure \ref{fig:dataset_format} illustrates the structure of our dataset. Sensor data were collected using the Robot Operating System (ROS), with ROS2 \citep{Macenski2022RobotOS} for the $\alpha$ device and ROS1 \citep{Quigley2009ROSAO} for other devices. To accommodate users who do not use ROS, we provide the bag files alongside raw data in various formats: human-readable text, JPEG images, and binary LiDAR point cloud files. Additionally, we include tools for visualizing point cloud data from these binary files. The raw data has been compressed to enhance accessibility.
We also supply calibration results for multiple sensors across three devices, a ground truth map for the stair sequence, and ground truth poses for each sequence. These resources ensure the dataset is comprehensive and useful for various applications and analyses.

\subsubsection{Raw data formats:}

The raw data is organized and stored according to sensor type, with each category of sensor data housed in its respective folder. For IMU sensors, the recorded data is documented in a text file format (.txt), with each log entry on a separate line. Each line is prefixed by a timestamp, followed by the IMU sensor readings (roll, pitch, yaw, angular velocities, and linear accelerations). Camera data is stored separately for the left and right cameras, extracted from the compressed images within ROS bag files, and saved in JPEG format. These images are named based on the timestamps of their capture. LiDAR sensor data is stored in binary files (.bin), containing the range measurements for each frame. Velodyne data includes \textit{x, y, z, intensity, ring, time}; Ouster data includes \textit{x, y, z, intensity, t, reflectivity, ring, ambient, range}; and LiVOX data includes \textit{x, y, z, intensity, tag, line}. These files contain essential data channels such as Cartesian coordinates, temporal offsets, and the indices of the rings or lines, crucial for data integrity and analysis.

\subsubsection{ROS Bag Files:}
The ROS bag file format functions as an extensive repository, encompassing the complete sensor data and encapsulating all ROS topics originating from the respective distinct collection devices, as outlined in Table \ref{topic_specs}. This format guarantees the integration and preservation of the various streams of sensor information, facilitating seamless accessibility for ROS-based algorithms.

\subsubsection{Calibration Files} 
The outcomes of the multi-sensor calibration, as described in Section \ref{subsec:Sensor_Calibration} , are recorded in three YAML files, each corresponding to the respective acquisition devices.

\begin{table*}[t]
    \centering
    \caption{Sensor Data Descriptions and Specifications}
    \label{topic_specs}
    \resizebox{\textwidth}{!}{
      \begin{tabular}{llllV{\linewidth}l}
        \toprule 
        Device & Sensor & Message Type & Topic Name & Description & Frequency\tabularnewline
        \midrule 
        \multirow{4}{*}{$\alpha$} & LiDAR & sensor\_msgs/PointCloud2 & \texttt{/velodyne\_points} & Raw velodyne pointcloud with
        
        [\textit{x, y, z, intensity, ring, time}] & 10$\ensuremath{Hz}$\tabularnewline
        \cmidrule{2-6}
         & \multirow{2}{*}{Camera} & sensor\_msgs/CompressedImage & \texttt{/left\_camera/compressed} & Compressed RGB image & 10$\ensuremath{Hz}$\tabularnewline
         &  & sensor\_msgs/CompressedImage & \texttt{/right\_camera/compressed} & Compressed RGB image & 10$\ensuremath{Hz}$\tabularnewline
        \cmidrule{2-6}
         & IMU & sensor\_msgs/Imu & \texttt{/imu/data} & Raw IMU data from Xsens MTi-30 & 100$\ensuremath{Hz}$\tabularnewline
        \midrule 
        \multirow{7}{*}{$\beta$} & \multirow{4}{*}{LiDAR} & sensor\_msgs/PointCloud2 & \texttt{/ouster/points} & Raw Ouster pointcloud with [\textit{x, y, z, }
        
        \textit{intensity, t, reflectivity, ring, ambient, range}] & 10$\ensuremath{Hz}$\tabularnewline
         &  & sensor\_msgs/Image & \texttt{/ouster/reflec\_image} & Reflectivity data & 10$\ensuremath{Hz}$\tabularnewline
         &  & sensor\_msgs/Image & \texttt{/ouster/range\_image} & Depth image & 10$\ensuremath{Hz}$\tabularnewline
         &  & sensor\_msgs/Imu & \texttt{/ouster/imu} & Raw IMU data from InvenSense ICM-20948 & 100$\ensuremath{Hz}$\tabularnewline
        \cmidrule{2-6}
         & \multirow{2}{*}{Camera} & sensor\_msgs/CompressedImage & \texttt{/left\_camera/image/compressed} & Compressed RGB image & 10$\ensuremath{Hz}$\tabularnewline
         &  & sensor\_msgs/CompressedImage & \texttt{/right\_camera/image/compressed} & Compressed RGB image & 10$\ensuremath{Hz}$\tabularnewline
        \cmidrule{2-6}
         & IMU & sensor\_msgs/Imu & \texttt{/imu/data} & Raw IMU data from Xsens MTi-30 & 100$\ensuremath{Hz}$\tabularnewline
        \midrule 
        \multirow{5}{*}{$\gamma$} & \multirow{2}{*}{LiDAR} & livox\_ros\_driver/CustomMsg & \texttt{/livox/lidar} & Livox defined cloudpoint data with
        
        [\textit{x, y, z, intensity, tag, line}] & 10$\ensuremath{Hz}$\tabularnewline
         &  & sensor\_msgs/Imu & \texttt{/livox/imu} & Raw IMU data from BMI088 & 200$\ensuremath{Hz}$\tabularnewline
        \cmidrule{2-6}
         & \multirow{2}{*}{Camera} & sensor\_msgs/CompressedImage & \texttt{/left\_camera/image/compressed} & Compressed RGB image & 10$\ensuremath{Hz}$\tabularnewline
         &  & sensor\_msgs/CompressedImage & \texttt{/right\_camera/image/compressed} & Compressed RGB image & 10$\ensuremath{Hz}$\tabularnewline
        \cmidrule{2-6}
         & IMU & sensor\_msgs/Imu & \texttt{/imu/data} & Raw IMU data from Xsens MTi-30 & 100$\ensuremath{Hz}$\tabularnewline
        \bottomrule
        \end{tabular}}
    \end{table*}

\begin{figure*}[t]
    \centering
    \subfloat[Devices for generating GT trajectories and maps: CHCNAV CG$610$, Vicon Vero 2.2, Leica Nova  MS$60$, and Leica RTC$360$.]{
          \includegraphics[width=0.9\linewidth]{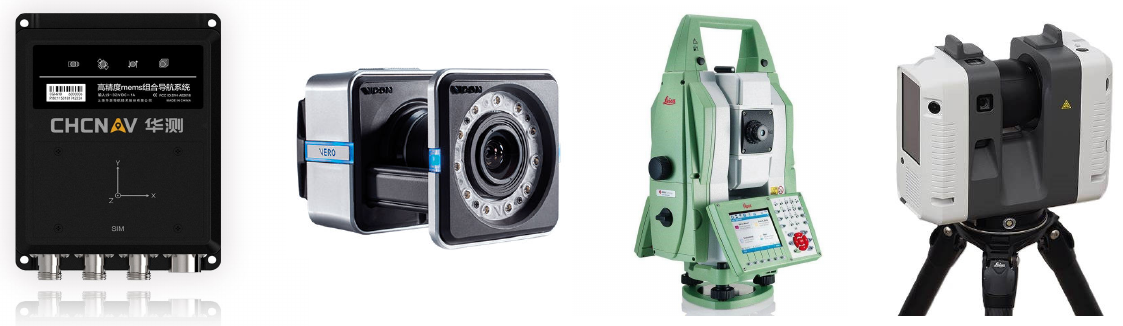}
      \label{fig:GT_devices}}
  
    \subfloat[GT RGB point cloud map of "Stairs" sequence.]{\includegraphics[width=0.6\textwidth]{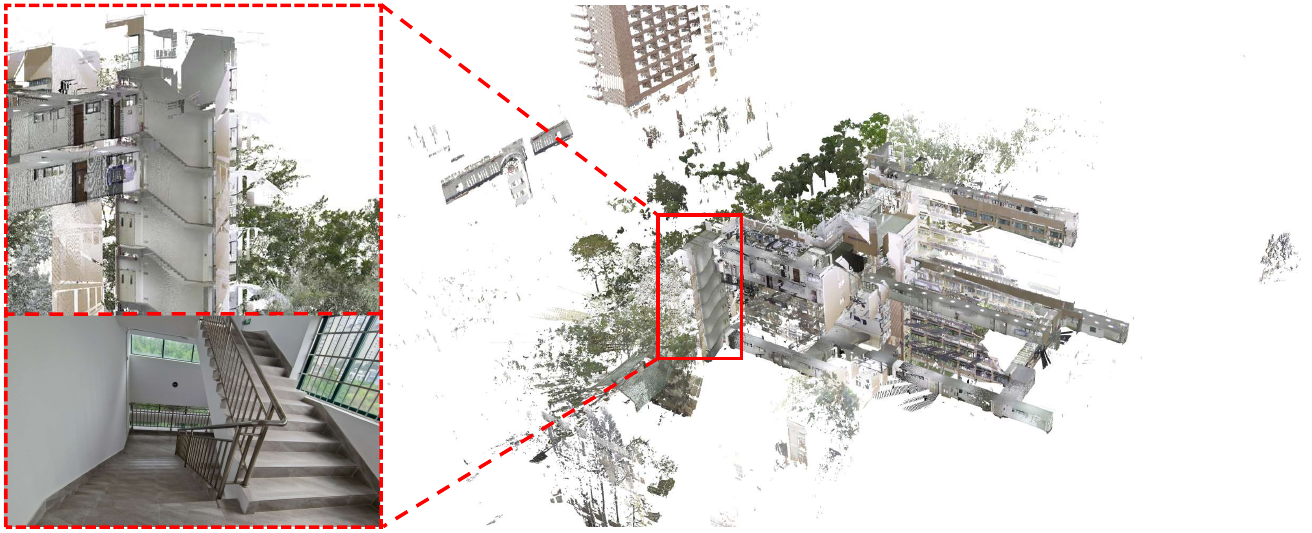}
    \label{fig:GTMap_color}}
    \subfloat[Groundtruth evaluation.]{\includegraphics[width=0.39\textwidth]{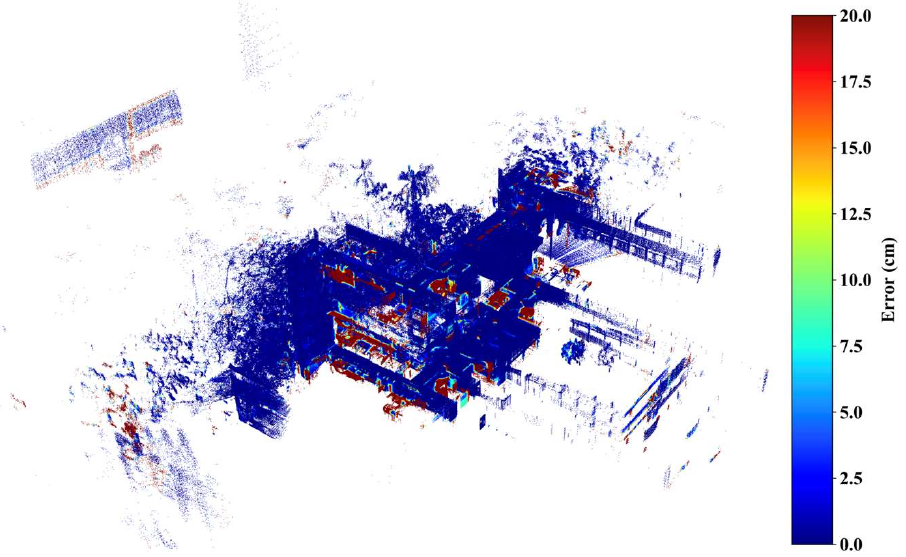}
    \label{fig:GTMap_error}}
    \caption{Groundtruth Generation.}
    \label{fig:GTMap}
  \end{figure*}

\subsection{Groundtruth Generation}
\subsubsection{LiDAR Point Cloud Maps Generation:}
For the stairs sequence, we employed a Leica RTC$360$ laser scanner to meticulously capture the details of the stairs on both sides of a building and the corridors across multiple floors with millimeter accuracy. 

\subsubsection{Groundtruth Poses:}

To capture accurate motion ground truth in various degraded environments of the GEODE dataset, it is essential to use appropriate equipment tailored to the specific scene characteristics, as illustrated in Figure \ref{fig:GT_devices}.
For outdoor sequences such as urban tunnels, bridges, inland waterways, and off-road settings, we use the RTK-INS device CHCNAV CG610 to obtain 6-DoF groundtruth poses. In the bridge and urban tunnel sequences, data collection relies solely on the $\alpha$ device. Hand-eye calibration \citep{Furrer2017EvaluationOC} is employed to refine the positioning results from the RTK-INS, ensuring accurate motion values.
In the inland waterway and off-road sequences, three acquisition devices are mounted on a rack constructed from aluminum profiles, which is then attached to a sailboat or UGV for data acquisition. The RTK-INS device's trajectory data is synchronized with the $\beta$ device using hand-eye calibration. Further processing with multi-LiDAR calibration \citep{Li2023JointIA} refines the motion values for the $\alpha$ and $\gamma$ devices.

\begin{table*}
    \begin{centering}
    \caption{Estimation Methods Comparison}
    \label{tab:estimation_method_comparison}
    \par\end{centering}
    \begin{centering}
    \resizebox{\textwidth}{!}{
        \begin{tabular}{llllll}
            \toprule 
            Method & Multisensor fusion & Degeneracy detection(\Circle) and mitigation(\CIRCLE) & Outlier rejection & Adaptive capabilities & Failure detection and recovery\tabularnewline
            \midrule
            FAST-LIO2 & LiDAR+IMU & \XSolidBrush{} & \XSolidBrush{} & \XSolidBrush{} & \XSolidBrush{}\tabularnewline
            \midrule 
            LIO-SAM & LiDAR+IMU & \makecell[l]{\Circle Optimization-based method\\ \CIRCLE Solution remapping techniques} & \XSolidBrush{} & \XSolidBrush{} & \XSolidBrush{}\tabularnewline
            \midrule 
            DLIO & LiDAR+IMU & \XSolidBrush{} & \XSolidBrush{} & \XSolidBrush{} & \XSolidBrush{}\tabularnewline
            \midrule 
            COIN-LIO & LiDAR+IMU+Intensity & \makecell[l]{\Circle Geometric method\\ \CIRCLE Geometrically complementary patch selection} & \XSolidBrush{} & \XSolidBrush{} & \XSolidBrush{}\tabularnewline
            \midrule 
            R3LIVE & LiDAR+IMU+Camera & \XSolidBrush{} & \XSolidBrush{} & \XSolidBrush{} & \XSolidBrush{}\tabularnewline
            \midrule 
            LVI-SAM & LiDAR+IMU+Camera & \makecell[l]{\Circle Optimization-based method\\ \CIRCLE Solution remapping techniques} & \XSolidBrush{} & \XSolidBrush{} & VIS failure detection\tabularnewline
            \midrule 
            RELEAD & LiDAR+IMU+Camera & \makecell[l]{\Circle Geometric method\\ \CIRCLE Constrained ESIKF update} & GNC-based pose outlier rejection & \XSolidBrush{} & \XSolidBrush{}\tabularnewline
            \bottomrule
            \end{tabular}}
    \par\end{centering}
    \end{table*}

For sequences on flat surfaces, 6-DoF ground truth is captured using the Vicon motion capture system. In metro tunnels, the Leica MS 60 is used to track prisms and obtain ground truth position. Temporal alignment between Leica MS60 total station measurements and sensor data is achieved using tool in FusionPortableV2 \citep{Wei2024FusionPortableV2AU}. Similar to the inland waterways sequence, some metro tunnel data is collected using three devices simultaneously. The trajectory data from the Leica MS60 is synchronized to the $\alpha$ device via hand-eye calibration, while the $\beta$ and $\gamma$ devices' ground truth is obtained through further multi-LiDAR calibration processing.

For stair sequences, the ground truth trajectory is generated using the PALoc SLAM method \citep{Hu2024PALocAS}, which estimates LiDAR poses within a pre-constructed map by the Leica RTC360 laser scanner Fig. \ref{fig:GTMap_color}. Figure \ref{fig:GTMap_error} illustrates maps constructed from the groundtruth poses generated by PALoc, highlighting errors from the groundtruth map in the sequence "stair\_bob". 
Due to the limited FOV of the Livox Avia LiDAR on the $\gamma$ device, the PALoc algorithm could not generate an accurate ground truth trajectory, even with a ground truth map available. Consequently, for this sequence, we advise users of this dataset to compare the map produced by their algorithm with the provided ground truth map to assess localization errors.

\subsection{Development Tools}
\label{sec:tool}

We offer a comprehensive development kit to assist users in effectively utilizing our dataset, particularly for evaluating algorithms implemented in Python.
Given the limited three degrees of freedom of the ground truth pose captured by the Leica MS 60 tracking prism in a metro tunnel, transforming these poses into the body coordinate systems of both the $\beta$ and $\gamma$ devices using multi-LiDAR calibration results can be challenging. To facilitate this process, we provide Python scripts that convert trajectory results from users' algorithms to the $\alpha$ device's body coordinate system for localization evaluation. Similarly, for the RTK-INS-based ground truth pose, which has six degrees of freedom, we offer scripts for converting user-generated trajectories to the $\beta$ device's coordinate system.

The development kit also includes scripts for calculating errors and visualizing trajectories, ensuring users have a comprehensive toolkit for algorithm assessment and validation.
Furthermore, we offer a comprehensive suite of C++ code, developed utilizing the software toolkit referenced in \cite{Morales2021TheUD}, designed to extract raw data from ROS bag files. Accompanying this, we also provide scripts tailored for visualizing LiDAR point clouds derived from binary data.

%% file: chapter/experiment.tex
\section{EVALUATION}  
\label{sec:experiment}
\runninghead{CHEN \textit{et~al.}}

This section delivers an extensive examination of cutting-edge LiDAR-based odometry methodologies, utilizing the GEODE dataset as a benchmark. The thorough assessment is designed to delineate the weekness of contemporary LiDAR odometry techniques and underscore the pivotal role that datasets akin to GEODE play in propelling robust algorithm forward. Our evaluation encompasses a rigorous testing of seven  state-of-the-art methods across the entire spectrum of available data sequences, as shown in Table \ref{tab:estimation_method_comparison}.

\begin{table*}
    \caption{Evaluation of ATE(M) of SLAM Systems on Dataset Sequences. \parbox{2em}{\centering\colorbox{bshade!110}{}} denotes the best result, and \parbox{2em}{\centering\colorbox{bshade!30}{}} denotes the second best result. \XSolidBrush{} denotes the algorithm breaking down or having a localization error larger than 100m. - denotes the SLAM algorithm is not adapted to this data.}
    \label{tab:rmse_on_sequences}
    \centering{}%
      \scalebox{0.72}{
      \begin{tabular}{cccccccccc}
        \toprule 
        \multicolumn{3}{c}{Sequence} & LIO-SAM & FAST-LIO2 & DLIO & COIN-LIO & LVI-SAM & R3LIVE & RELEAD\tabularnewline
        \midrule 
        \multirow{21}{*}{Offroad} & \multirow{7}{*}{$\alpha$} & 01 & 0.23 & \colorbox{bshade!110}{\makebox[4em][c]{\textbf{0.18}}} & \XSolidBrush{} & - & \XSolidBrush{} & \XSolidBrush{} & \colorbox{bshade!30}{\makebox[4em][c]{\textbf{0.22}}}\tabularnewline
         &  & 02 & \XSolidBrush{} & \colorbox{bshade!30}{\makebox[4em][c]{\textbf{0.39}}} & \XSolidBrush{} & - & \XSolidBrush{} & \XSolidBrush{} & \colorbox{bshade!110}{\makebox[4em][c]{\textbf{0.26}}}\tabularnewline
         &  & 03 & \XSolidBrush{} & \colorbox{bshade!30}{\makebox[4em][c]{\textbf{0.33}}} & \XSolidBrush{} & - & \XSolidBrush{} & \XSolidBrush{} & \colorbox{bshade!110}{\makebox[4em][c]{\textbf{0.19}}}\tabularnewline
         &  & 04 & \XSolidBrush{} & \colorbox{bshade!30}{\makebox[4em][c]{\textbf{0.43}}} & \XSolidBrush{} & - & \XSolidBrush{} & \XSolidBrush{} & \colorbox{bshade!110}{\makebox[4em][c]{\textbf{0.13}}}\tabularnewline
         &  & 05 & \XSolidBrush{} & \colorbox{bshade!30}{\makebox[4em][c]{\textbf{0.30}}} & \XSolidBrush{} & - & \XSolidBrush{} & \XSolidBrush{} & \colorbox{bshade!110}{\makebox[4em][c]{\textbf{0.25}}}\tabularnewline
         &  & 06 & \XSolidBrush{} & \colorbox{bshade!30}{\makebox[4em][c]{\textbf{0.22}}} & \XSolidBrush{} & - & \XSolidBrush{} & \XSolidBrush{} & \colorbox{bshade!110}{\makebox[4em][c]{\textbf{0.13}}}\tabularnewline
         &  & 07 & 0.49 & \colorbox{bshade!30}{\makebox[4em][c]{\textbf{0.26}}} & \XSolidBrush{} & - & \XSolidBrush{} & \XSolidBrush{} & \colorbox{bshade!110}{\makebox[4em][c]{\textbf{0.13}}}\tabularnewline
        \cmidrule{2-10}
         & \multirow{7}{*}{$\beta$} & 01 & 0.15 & \colorbox{bshade!30}{\makebox[4em][c]{\textbf{0.12}}} & 0.18 & \colorbox{bshade!30}{\makebox[4em][c]{\textbf{0.12}}} & 0.14 & 5.42 & \colorbox{bshade!110}{\makebox[4em][c]{\textbf{0.10}}}\tabularnewline
         &  & 02 & \XSolidBrush{} & 0.31 & 0.22 & \colorbox{bshade!30}{\makebox[4em][c]{\textbf{0.13}}} & 0.20 & 20.95 & \colorbox{bshade!110}{\makebox[4em][c]{\textbf{0.11}}}\tabularnewline
         &  & 03 & \XSolidBrush{} & 0.34 & 0.18 & \colorbox{bshade!110}{\makebox[4em][c]{\textbf{0.12}}} & 2.76 & 7.68 & \colorbox{bshade!30}{\makebox[4em][c]{\textbf{0.13}}}\tabularnewline
         &  & 04 & \XSolidBrush{} & 0.14 & 0.23 & \colorbox{bshade!30}{\makebox[4em][c]{\textbf{0.13}}} & \XSolidBrush{} & 37.79 & \colorbox{bshade!110}{\makebox[4em][c]{\textbf{0.11}}}\tabularnewline
         &  & 05 & \XSolidBrush{} & 0.19 & 0.37 & \colorbox{bshade!30}{\makebox[4em][c]{\textbf{0.16}}} & 8.33 & 42.43 & \colorbox{bshade!110}{\makebox[4em][c]{\textbf{0.13}}}\tabularnewline
         &  & 06 & 3.15 & \colorbox{bshade!30}{\makebox[4em][c]{\textbf{0.15}}} & 0.21 & \colorbox{bshade!30}{\makebox[4em][c]{\textbf{0.15}}} & \XSolidBrush{} & 4.78 & \colorbox{bshade!110}{\makebox[4em][c]{\textbf{0.11}}}\tabularnewline
         &  & 07 & 0.19 & \colorbox{bshade!30}{\makebox[4em][c]{\textbf{0.11}}} & 0.15 & \colorbox{bshade!30}{\makebox[4em][c]{\textbf{0.11}}} & 19.19 & 3.77 & \colorbox{bshade!110}{\makebox[4em][c]{\textbf{0.10}}}\tabularnewline
        \cmidrule{2-10}
         & \multirow{7}{*}{$\gamma$} & 01 & \XSolidBrush{} & \colorbox{bshade!30}{\makebox[4em][c]{\textbf{0.09}}} & 1.19 & - & 2.43 & \XSolidBrush{} & \colorbox{bshade!110}{\makebox[4em][c]{\textbf{0.08}}}\tabularnewline
         &  & 02 & \XSolidBrush{} & \colorbox{bshade!110}{\makebox[4em][c]{\textbf{0.11}}} & 8.81 & - & 6.40 & \XSolidBrush{} & \colorbox{bshade!30}{\makebox[4em][c]{\textbf{0.12}}}\tabularnewline
         &  & 03 & \XSolidBrush{} & \colorbox{bshade!110}{\makebox[4em][c]{\textbf{0.09}}} & \colorbox{bshade!30}{\makebox[4em][c]{\textbf{0.29}}} & - & 1.14 & \XSolidBrush{} & \colorbox{bshade!110}{\makebox[4em][c]{\textbf{0.09}}}\tabularnewline
         &  & 04 & \XSolidBrush{} & \XSolidBrush{} & 3.55 & - & \colorbox{bshade!30}{\makebox[4em][c]{\textbf{2.61}}} & \XSolidBrush{} & \colorbox{bshade!110}{\makebox[4em][c]{\textbf{0.15}}}\tabularnewline
         &  & 05 & \XSolidBrush{} & \colorbox{bshade!110}{\makebox[4em][c]{\textbf{0.13}}} & 17.58 & - & 1.60 & \XSolidBrush{} & \colorbox{bshade!30}{\makebox[4em][c]{\textbf{0.14}}}\tabularnewline
         &  & 06 & \XSolidBrush{} & \colorbox{bshade!30}{\makebox[4em][c]{\textbf{0.12}}} & 0.87 & - & 27.04 & \XSolidBrush{} & \colorbox{bshade!110}{\makebox[4em][c]{\textbf{0.11}}}\tabularnewline
         &  & 07 & \XSolidBrush{} & \colorbox{bshade!30}{\makebox[4em][c]{\textbf{0.16}}} & 5.71 & - & 1.11 & 5.65 & \colorbox{bshade!110}{\makebox[4em][c]{\textbf{0.15}}}\tabularnewline
        \midrule 
        \multirow{9}{*}{Inland Waterways} & \multirow{3}{*}{$\alpha$} & Short & \XSolidBrush{} & 10.27 & \colorbox{bshade!110}{\makebox[4em][c]{\textbf{2.59}}} & - & \XSolidBrush{} & \colorbox{bshade!30}{\makebox[4em][c]{\textbf{4.80}}} & 8.5\tabularnewline
         &  & Medium & \XSolidBrush{} & \colorbox{bshade!30}{\makebox[4em][c]{\textbf{19.69}}} & \colorbox{bshade!110}{\makebox[4em][c]{\textbf{13.25}}} & - & \XSolidBrush{} & 22.77 & 22.81\tabularnewline
         &  & Long & \XSolidBrush{} & 70.26 & \colorbox{bshade!30}{\makebox[4em][c]{\textbf{68.04}}} & - & \XSolidBrush{} & \XSolidBrush{} & \colorbox{bshade!110}{\makebox[4em][c]{\textbf{67.40}}}\tabularnewline
        \cmidrule{2-10}
         & \multirow{3}{*}{$\beta$} & Short & \XSolidBrush{} & 0.29 & \colorbox{bshade!30}{\makebox[4em][c]{\textbf{0.15}}} & \colorbox{bshade!110}{\makebox[4em][c]{\textbf{0.13}}} & \XSolidBrush{} & 3.53 & 0.21\tabularnewline
         &  & Medium & \XSolidBrush{} & \colorbox{bshade!30}{\makebox[4em][c]{\textbf{0.57}}} & 0.72 & \colorbox{bshade!110}{\makebox[4em][c]{\textbf{0.49}}} & \XSolidBrush{} & 47.68 & 0.78\tabularnewline
         &  & Long & \XSolidBrush{} & 0.99 & \colorbox{bshade!30}{\makebox[4em][c]{\textbf{0.82}}} & \colorbox{bshade!110}{\makebox[4em][c]{\textbf{0.78}}} & \XSolidBrush{} & \XSolidBrush{} & 7.88\tabularnewline
        \cmidrule{2-10}
         & \multirow{3}{*}{$\gamma$} & Short & \XSolidBrush{} & \colorbox{bshade!110}{\makebox[4em][c]{\textbf{0.24}}} & 0.39 & - & 23.81 & 0.35 & \colorbox{bshade!30}{\makebox[4em][c]{\textbf{0.28}}}\tabularnewline
         &  & Medium & \colorbox{bshade!110}{\makebox[4em][c]{\textbf{1.14}}} & \colorbox{bshade!30}{\makebox[4em][c]{\textbf{1.82}}} & 2.75 & - & \XSolidBrush{} & 5.73 & 10.84\tabularnewline
         &  & Long & \XSolidBrush{} & 6.90 & \colorbox{bshade!110}{\makebox[4em][c]{\textbf{5.34}}} & - & \XSolidBrush{} & 12.66 & \colorbox{bshade!30}{\makebox[4em][c]{\textbf{5.59}}}\tabularnewline
        \midrule 
        \multirow{23}{*}{Metro Tunnels} & \multirow{4}{*}{$\alpha$} & Tunneling2 & \XSolidBrush{} & \XSolidBrush{} & \XSolidBrush{} & - & \XSolidBrush{} & \XSolidBrush{} & \XSolidBrush{}\tabularnewline
         &  & Tunneling3 & \colorbox{bshade!110}{\makebox[4em][c]{\textbf{0.18}}} & 0.21 & \colorbox{bshade!110}{\makebox[4em][c]{\textbf{0.18}}} & - & \XSolidBrush{} & 0.59 & \colorbox{bshade!30}{\makebox[4em][c]{\textbf{0.19}}}\tabularnewline
         &  & Tunneling4 & \colorbox{bshade!30}{\makebox[4em][c]{\textbf{0.18}}} & 0.24 & \colorbox{bshade!110}{\makebox[4em][c]{\textbf{0.14}}} & - & \XSolidBrush{} & 0.25 & 0.19\tabularnewline
         &  & Tunneling5 & \colorbox{bshade!30}{\makebox[4em][c]{\textbf{0.16}}} & 0.19 & \colorbox{bshade!110}{\makebox[4em][c]{\textbf{0.13}}} & - & 0.30 & 11.44 & 0.17\tabularnewline
        \cmidrule{2-10}
         & \multirow{8}{*}{$\beta$} & Shield7 & \XSolidBrush{} & \XSolidBrush{} & \XSolidBrush{} & \XSolidBrush{} & \XSolidBrush{} & \XSolidBrush{} & \XSolidBrush{}\tabularnewline
         &  & Shield8 & \XSolidBrush{} & \XSolidBrush{} & \XSolidBrush{} & \XSolidBrush{} & \XSolidBrush{} & \XSolidBrush{} & \XSolidBrush{}\tabularnewline
         &  & Shield9 & \XSolidBrush{} & \XSolidBrush{} & \XSolidBrush{} & \XSolidBrush{} & \XSolidBrush{} & \XSolidBrush{} & \colorbox{bshade!110}{\makebox[4em][c]{\textbf{12.69}}}\tabularnewline
         &  & Shield10 & \XSolidBrush{} & \XSolidBrush{} & \XSolidBrush{} & \XSolidBrush{} & \XSolidBrush{} & \XSolidBrush{} & \colorbox{bshade!110}{\makebox[4em][c]{\textbf{79.63}}}\tabularnewline
         &  & Tunneling2 & \colorbox{bshade!30}{\makebox[4em][c]{\textbf{0.13}}} & 0.14 & \colorbox{bshade!110}{\makebox[4em][c]{\textbf{0.11}}} & 0.14 & 0.14 & 0.14 & \colorbox{bshade!30}{\makebox[4em][c]{\textbf{0.13}}}\tabularnewline
         &  & Tunneling3 & 0.17 & 0.17 & \colorbox{bshade!110}{\makebox[4em][c]{\textbf{0.14}}} & 0.23 & 0.37 & 0.20 & \colorbox{bshade!30}{\makebox[4em][c]{\textbf{0.16}}}\tabularnewline
         &  & Tunneling4 & 0.18 & 0.17 & \colorbox{bshade!110}{\makebox[4em][c]{\textbf{0.15}}} & \colorbox{bshade!30}{\makebox[4em][c]{\textbf{0.16}}} & 0.18 & 0.25 & 0.17\tabularnewline
         &  & Tunneling5 & 0.13 & \colorbox{bshade!30}{\makebox[4em][c]{\textbf{0.12}}} & \colorbox{bshade!110}{\makebox[4em][c]{\textbf{0.11}}} & 3.84 & 0.22 & 0.26 & 0.13\tabularnewline
        \cmidrule{2-10}
         & \multirow{11}{*}{$\gamma$} & Tunneling1 & \colorbox{bshade!110}{\makebox[4em][c]{\textbf{0.33}}} & 1.16 & 8.51 & - & 35.40 & 2.83 & \colorbox{bshade!30}{\makebox[4em][c]{\textbf{0.37}}}\tabularnewline
         &  & Tunneling2 & 3.84 & \colorbox{bshade!110}{\makebox[4em][c]{\textbf{1.88}}} & 15.43 & - & \colorbox{bshade!30}{\makebox[4em][c]{\textbf{2.10}}} & 86.51 & 2.32\tabularnewline
         &  & Tunneling3 & 0.28 & \colorbox{bshade!110}{\makebox[4em][c]{\textbf{0.19}}} & 2.63 & - & 0.24 & 63.05 & \colorbox{bshade!30}{\makebox[4em][c]{\textbf{0.22}}}\tabularnewline
         &  & Tunneling4 & 1.28 & \colorbox{bshade!110}{\makebox[4em][c]{\textbf{0.15}}} & 5.74 & - & \colorbox{bshade!30}{\makebox[4em][c]{\textbf{0.35}}} & 57.46 & \colorbox{bshade!110}{\makebox[4em][c]{\textbf{0.15}}}\tabularnewline
         &  & Tunneling5 & 14.77 & \colorbox{bshade!30}{\makebox[4em][c]{\textbf{0.27}}} & 2.48 & - & \colorbox{bshade!110}{\makebox[4em][c]{\textbf{0.17}}} & 1.30 & 0.35\tabularnewline
         &  & Shield1 & \XSolidBrush{} & \XSolidBrush{} & \XSolidBrush{} & - & \XSolidBrush{} & \XSolidBrush{} & \XSolidBrush{}\tabularnewline
         &  & Shield2 & \XSolidBrush{} & \XSolidBrush{} & \XSolidBrush{} & - & \XSolidBrush{} & \XSolidBrush{} & \XSolidBrush{}\tabularnewline
         &  & Shield3 & \XSolidBrush{} & \XSolidBrush{} & \XSolidBrush{} & - & \XSolidBrush{} & \XSolidBrush{} & \colorbox{bshade!110}{\makebox[4em][c]{\textbf{3.20}}}\tabularnewline
         &  & Shield4 & \XSolidBrush{} & \XSolidBrush{} & \XSolidBrush{} & - & \XSolidBrush{} & \XSolidBrush{} & \XSolidBrush{}\tabularnewline
         &  & Shield5 & \XSolidBrush{} & \XSolidBrush{} & \XSolidBrush{} & - & \XSolidBrush{} & \XSolidBrush{} & \colorbox{bshade!110}{\makebox[4em][c]{\textbf{4.49}}}\tabularnewline
         &  & Shield6 & \XSolidBrush{} & \XSolidBrush{} & \XSolidBrush{} & - & \XSolidBrush{} & \XSolidBrush{} & \colorbox{bshade!110}{\makebox[4em][c]{\textbf{1.43}}}\tabularnewline
        \midrule 
        \multirow{3}{*}{Stairs} & $\alpha$ &  & 6.30 & 4.69 & 4.89 & - & \XSolidBrush{} & \colorbox{bshade!30}{\makebox[4em][c]{\textbf{4.54}}} & \colorbox{bshade!110}{\makebox[4em][c]{\textbf{0.57}}}\tabularnewline
         & $\beta$ &  & 4.19 & \colorbox{bshade!30}{\makebox[4em][c]{\textbf{0.38}}} & 0.41 & 0.45 & \XSolidBrush{} & \XSolidBrush{} & \colorbox{bshade!110}{\makebox[4em][c]{\textbf{0.21}}} \tabularnewline
         & $\gamma$ &  & \XSolidBrush{} & \XSolidBrush{} & \XSolidBrush{} & - & \XSolidBrush{} & \XSolidBrush{} & \XSolidBrush{} \tabularnewline
        \midrule 
        \multirow{3}{*}{Bridges} & \multirow{3}{*}{$\alpha$} & 01 & \XSolidBrush{} & \XSolidBrush{} & \XSolidBrush{} & - & \XSolidBrush{} & \XSolidBrush{} & \XSolidBrush{}\tabularnewline
         &  & 02 & \XSolidBrush{} & \XSolidBrush{} & \XSolidBrush{} & - & \XSolidBrush{} & \XSolidBrush{} & \XSolidBrush{}\tabularnewline
         &  & 03 & \XSolidBrush{} & \XSolidBrush{} & \XSolidBrush{} & - & \XSolidBrush{} & \XSolidBrush{} & \XSolidBrush{}\tabularnewline
        \midrule 
        \multirow{3}{*}{Urban Tunnels} & \multirow{3}{*}{$\alpha$} & 01 & \XSolidBrush{} & \XSolidBrush{} & \XSolidBrush{} & - & \XSolidBrush{} & \XSolidBrush{} & \colorbox{bshade!110}{\makebox[4em][c]{\textbf{40.50}}}\tabularnewline
         &  & 02 & \XSolidBrush{} & \XSolidBrush{} & \XSolidBrush{} & - & \XSolidBrush{} & \XSolidBrush{} & \XSolidBrush{}\tabularnewline
         &  & 03 & \XSolidBrush{} & \XSolidBrush{} & \XSolidBrush{} & - & \XSolidBrush{} & \XSolidBrush{} & \XSolidBrush{}\tabularnewline
        \midrule 
        \multirow{2}{*}{Flat Ground} & \multirow{2}{*}{$\gamma$} & smooth & \XSolidBrush{} & \XSolidBrush{} & \XSolidBrush{} & - & \XSolidBrush{} & \XSolidBrush{} & \colorbox{bshade!110}{\makebox[4em][c]{\textbf{0.26}}}\tabularnewline
         &  & aggressive & \XSolidBrush{} & \XSolidBrush{} & \XSolidBrush{} & - & \XSolidBrush{} & \XSolidBrush{} & \colorbox{bshade!110}{\makebox[4em][c]{\textbf{1.17}}}\tabularnewline
        \bottomrule
        \end{tabular}
        }
    \end{table*}

\subsection{Baselines}
FAST-LIO2 \citep{Xu2021FASTLIO2FD} offers a dependable solution for LiDAR-inertial odometry by fusing IMU data with LiDAR feature points through an iterative extended Kalman filter. It enables efficient navigation in challenging environments characterized by rapid motion, noise, or clutter, where performance degradation may occur.

LIO-SAM \citep{Shan2020LIOSAMTL} is a package that performs tight coupling of IMU and LiDAR data, incorporating degeneracy factor and solution remapping techniques to address geometric degeneracy issues.

DLIO \citep{Chen2022DirectLO} is a newly introduced odometry that construct continuous-time trajectories for precise motion correction. It introduces several enhancements over its predecessor, DLO \citep{Chen2021DirectLO}, initially developed for the DARPA Challenge.

COIN-LIO \citep{Pfreundschuh2023COINLIOCI} proposes an pipeline leverage intensity as an additional modality to improve the robustness of LiDAR-inertial odometry in geometrically degenerate scenarios.

R3LIVE \citep{Lin2021R3LIVEAR} is a fusion framework that combines LiDAR, inertial, and visual sensors to achieve reliable and precise state estimation in challenging environments with geometric degeneracy.

LVI-SAM \citep{Shan2021LVISAMTL} is a system that effectively tackles degeneration by fusing lidar and visual data in odometry, capitalizing on the strengths of each modality.

RELEAD \citep{Chen2024RELEADRL} integrates degeneracy detection and mitigation modules with a failure-tolerant multi-sensor fusion framework for maintaining well-constrained system states in LiDAR-degenerate scenarios.


\subsection{Accuracy Evaluation}
Table \ref{tab:rmse_on_sequences} summarizes the Absolute Trajectory Errors (ATE) observed in real-world geometrically degraded scenarios for seven SLAM algorithms. The results highlight key areas where existing SLAM systems can be improved. Despite being designed for robust localization, most algorithms still face challenges in geometrically degenerate environments. Three primary limitations of current LiDAR-centric SLAM algorithms contribute to this struggle.

\subsubsection{Irrobust to geometrical degradation:}

Current open-source LIO and LVIO systems exhibit limited robustness when dealing with geometric degradation. Feature-based approaches, such as LIO-SAM and LVI-SAM, align point clouds using only the most critical points. However, these methods require a computationally expensive feature extraction step that may inadvertently discard valuable data, potentially compromising downstream registration quality.
In scenarios involving geometric degradation, features useful for localization tend to be sparse. Consequently, feature extraction can result in significant portions of geometric information being ignored, leading to performance degradation. For instance, while direct methods like FAST-LIO and DLIO can achieve proper localization in off-road sequences, LIO-SAM with feature extraction shows substantial drift.
Dense methods use complete point clouds to avoid errors in off-road sequences where limited information from objects like electric poles still exists. These methods mitigate some of the issues associated with sparse feature extraction by leveraging the entire point cloud. 
Failures are common in environments with identical geometry, such as tunnels and bridges, where the lack of distinct features causes all methods to fail at localization. This challenge underscores the need for more robust approaches that can handle such uniform environments effectively.

\subsubsection{Without adaptive capabilities:}
\begin{figure}[t]
  \centering
  \subfloat[Point cloud sampling.]{\includegraphics[width=0.245\textwidth]{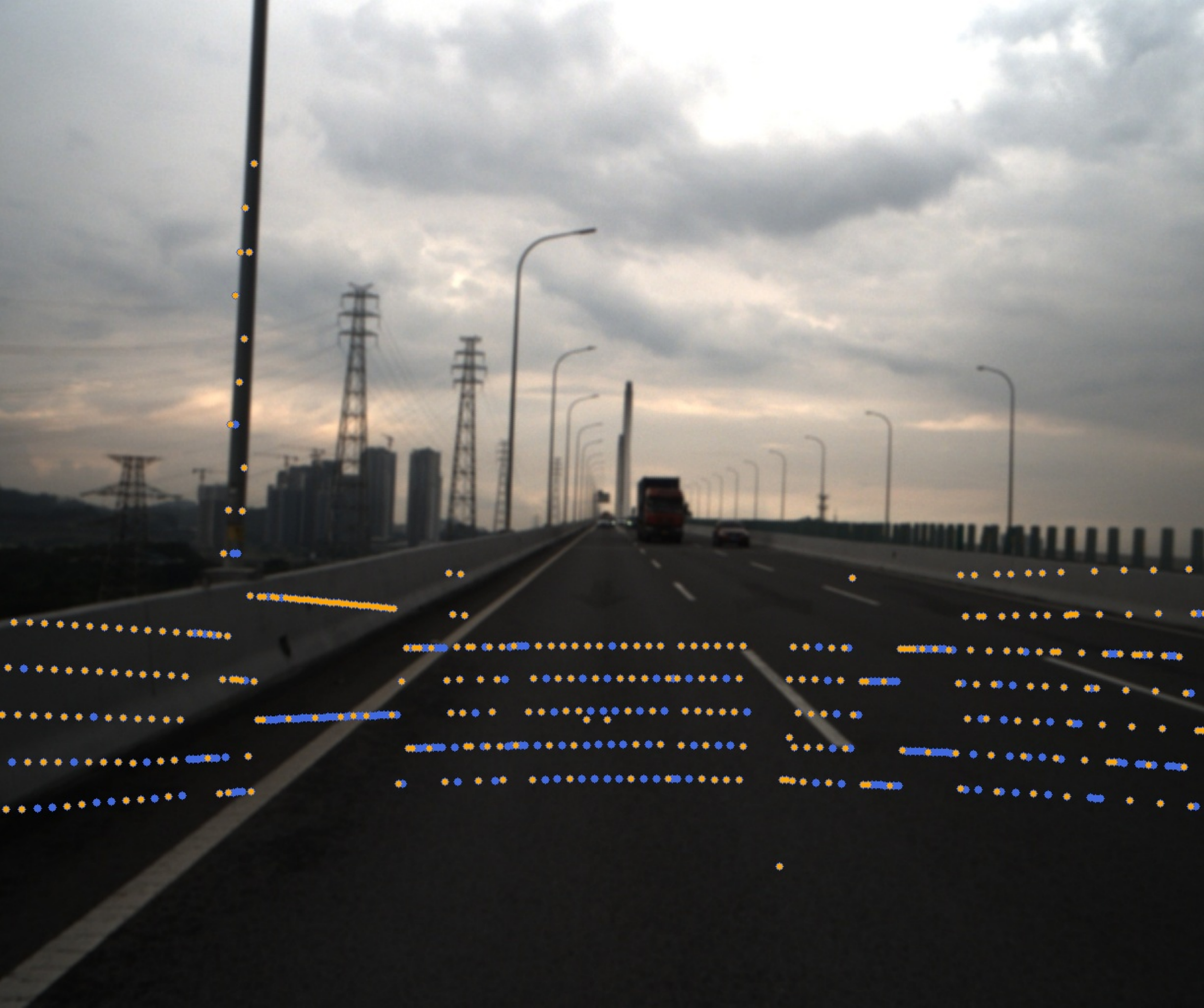}
  \label{fig:adaptive_feature_tracking_lidar}}
  \subfloat[Visaul feature point tracking.]{\includegraphics[width=0.245\textwidth]{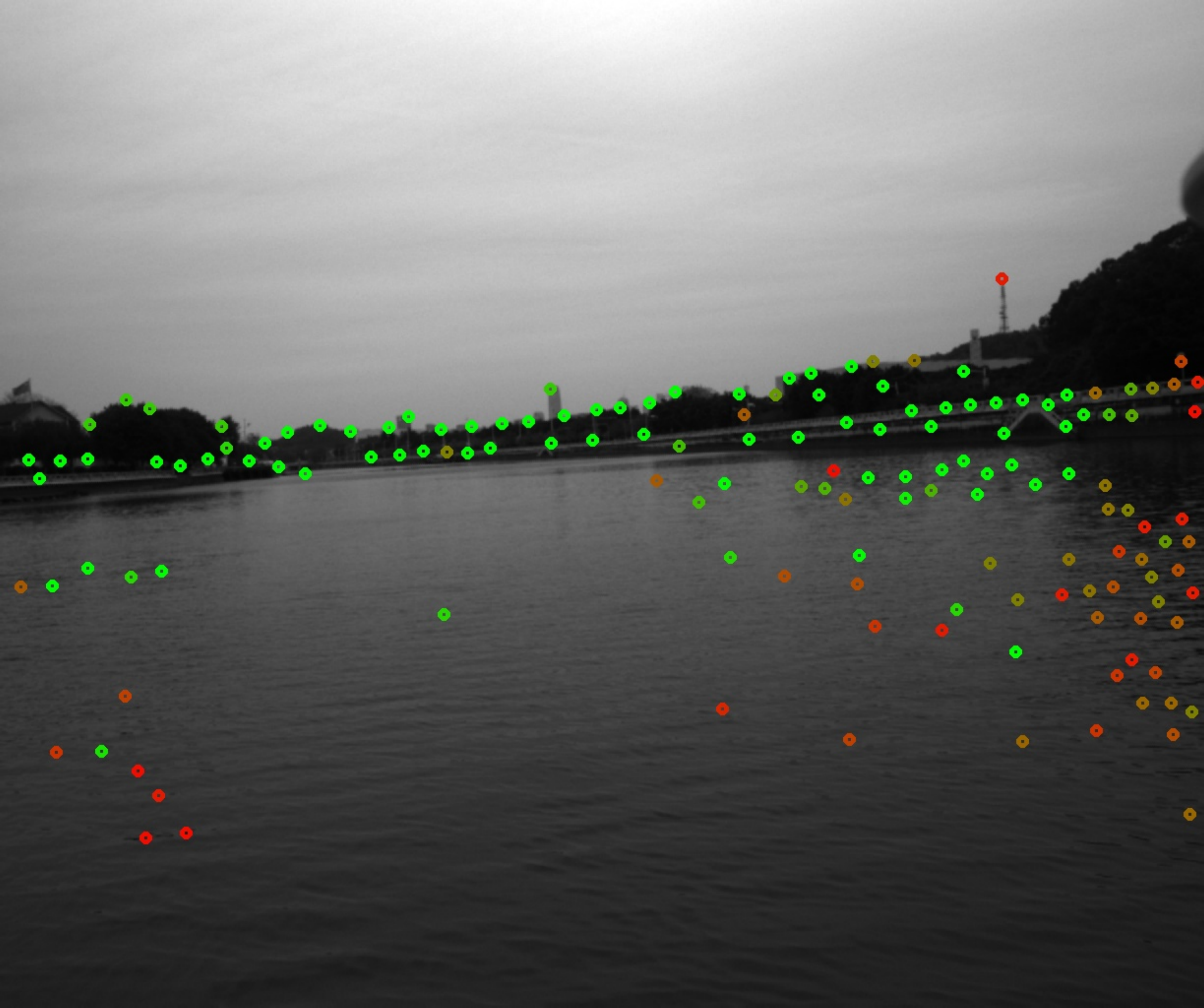}
  \label{fig:adaptive_feature_tracking_visual}}
  \caption{Lack of adaptive feature tracking capability.}
  \label{fig:adaptive_feature_tracking}
\end{figure}

A significant limitation of current SLAM methods is their inability to actively select the most informative measurements to adapt to various environments. This issue is evident in two main aspects. 
Firstly, single-sensor odometry lacks the capability for adaptive feature tracking. For example, all LIO methods failed to utilize the steel cables of the Cable Stayed Bridge and street lights as key localization features in the bridge sequence. Consequently, points that could be accurately located in the degraded direction (represented by orange points in Figure \ref{fig:adaptive_feature_tracking_lidar}, as assessed by the RMS algorithm \citep{Petrek2023RMSRP}) are overshadowed by repetitive, non-contributing points (blue dots in Figure \ref{fig:adaptive_feature_tracking_lidar}), leading to significant drift. Similarly, the VIO module in multi-sensor fusion methods such as LVI-SAM, R3LIVE and RELEAD does not adequately reduce the weight of feature points on water surfaces in inland waterway sequences, resulting in visual localization drift, as depicted in Figure \ref{fig:adaptive_feature_tracking}.

Secondly, multi-sensor fusion algorithms often fail to dynamically select complementary sensors to maintain uninterrupted localization. In tunnel sequences, which are characterized by varying exposure levels and featureless surroundings, algorithms designed to combine LiDAR and visual sensors should leverage visual data in geometrically degraded contexts while ignoring it under over-exposure conditions. Similarly, environments such as inland waterways and flat ground sequences present challenges for algorithms to quickly identify sensor degeneration and switch to alternative modalities for localization. However, current multi-sensor fusion algorithms have consistently underperformed in such scenarios, highlighting their limitations.

\subsubsection{Lack of Failure Detection and Recovery:}

Experimental results indicate that current open-source algorithms lack a dynamic adaptive architecture to effectively address algorithmic failures. 
This deficiency manifests in two primary ways: the failure to accurately detect sensor failure states and the lack of mechanisms to recover from these failures, resulting in significant challenges when coping with real-world degradation.

For instance, the degradation detection modules in LIO-SAM and LVI-SAM fail to correctly identify the direction of degradation in point cloud alignment, relying on auxiliary state estimation that is highly susceptible to noise. Similarly, R3LIVE lacks mechanisms for degeneracy detection and outlier measurement rejection, leading to failures in self-similar areas (affecting LiDAR) and under poor lighting conditions (affecting vision).
Although RELEAD has a mechanism to reject visual odometry outliers, its loosely coupled algorithm structure does not fully utilize multi-sensor information to overcome degraded scenarios, especially when visual localization is continuously unavailable. Consequently, despite achieving better localization results in tunnels compared to other algorithms, RELEAD is still far from being robustly usable.

Moreover, current odometry algorithms lack the capability to recover localization after submodule failures. Despite LVI-SAM's claim of using a multi-sensor subsystem to reinitialize failed subsystems, it failed to reorient the LiDAR inertial odometers after experiencing geometric degradation in tests.
These observations underscore the need for more robust and adaptive algorithms capable of handling sensor failures and geometric degradation in diverse, challenging environments.

\subsection{Future Research Directions}

To achieve accurate odometry estimates in geometrically degenerated scenarios, resilient methods capable of dynamic reconfiguration are required. 
This involves several aspects, such as selecting residuals, adding constraints \citep{Tuna2024,tuna2024informedconstrainedalignedfield}, online adaptive tuning of parameters, and even switching algorithmic modules. 
For example, treating the residuals in different degrees of freedom independently for LIO methods can improve applicability by reducing the need for domain-specific tuning.

For visual odometry that assists LiDAR odometry, challenges such as texture repetition, lighting variations, and occlusion must be addressed, and outlying points should be automatically rejected \citep{Bai2021DegenerationAwareOM}. In multi-sensor fusion, it is crucial to assign appropriate weights to different measurements to enhance the system's robustness. Determining the reliability of each sensor source in various directions is essential in a multi-source fusion system, highlighting the significant research value in recognizing degradation scenarios.

Another important area of research is how to recover the algorithmic module after a failure. Extending the robustness of LiDAR SLAM to geometrically degenerate scenarios requires redundancy and resourcefulness in information to constrain state optimization in the degenerated direction. This should be achieved through means beyond merely adding more sensors. It is necessary to regain performance in the face of degraded sensing or environmental changes.